\documentclass[runningheads]{llncs}

\usepackage{accv}

\usepackage{accvabbrv}

\usepackage{graphicx}
\usepackage{booktabs}
\usepackage{kotex}
\usepackage[ruled]{algorithm2e}
\usepackage{mathtools}
\usepackage{makecell}
\usepackage{multirow}
\usepackage{subcaption}
\usepackage{wrapfig}
\usepackage{float}

\usepackage[accsupp]{axessibility}  % Improves PDF readability for those with disabilities.

\usepackage{hyperref}

\begin{document}

% ---------------------------------------------------------------
% TODO REVIEW: Replace with your title
\title{Diffusion Model Compression\\for Image-to-Image Translation}

% TODO REVIEW: If the paper title is too long for the running head, you can set
% an abbreviated paper title here. If not, comment out.
\titlerunning{ID-Compression}

% TODO FINAL: Replace with your author list. 
% Include the authors' OCRID for the camera-ready version, if at all possible.
\author{Geonung Kim \and
Beomsu Kim \and
Eunhyeok Park \and
Sunghyun Cho}

% TODO FINAL: Replace with an abbreviated list of authors.
\authorrunning{G.~Kim et al.}
% First names are abbreviated in the running head.
% If there are more than two authors, 'et al.' is used.

% TODO FINAL: Replace with your institution list.
\institute{POSTECH\\
\email{\{k2woong92,qjatn0120,eh.park,s.cho\}@postech.ac.kr}\\
\url{https://kimgeonung.github.io/id-compression}
}

\maketitle

\def\MethodName{Unknow} 

\newcommand{\Eq}[1]  {Eq.\ (#1)}
\newcommand{\Eqs}[1] {Eqs.\ (#1)}
\newcommand{\Fig}[1] {Fig.\ #1}
\newcommand{\Figs}[1]{Figs.\ #1}
\newcommand{\Tbl}[1]  {Tab.\ #1}
\newcommand{\Tbls}[1] {Tabs.\ #1}
\newcommand{\Sec}[1] {Sec.\ #1}
\newcommand{\SSec}[1] {Sec.\ #1}
\newcommand{\Secs}[1] {Secs.\ #1}
\newcommand{\Alg}[1] {Alg.\ #1}

\newcommand{\setone}[1] {\left\{ #1 \right\}} % math set notation { a }
\newcommand{\settwo}[2] {\left\{ #1 \mid #2 \right\}} % math set notation { a | b}

\newcommand{\kkw}[1]{{\textcolor{magenta}{[KGU: #1]}}}
\newcommand{\sunghyun}[1]{{\textcolor[rgb]{0.6,0.0,0.6}{sunghyun: #1}}}
\newcommand{\eh}[1]{{\textcolor[rgb]{0.0,0.6,0.6}{eh: #1}}}

\newcommand{\bb}[1]{\textbf{\textit{#1}}}
\newcommand{\xx}{\textcolor{red}{XX}}
\newcommand{\fix}[1]{\textcolor{red}{#1}}
\newcommand{\rt}[1]{\textcolor{red}{#1}}

\newcommand\Myperm[2][^n]{\prescript{#1\mkern-2.5mu}{}P_{#2}}
\newcommand\Mycomb[2][^n]{\prescript{#1\mkern-0.5mu}{}C_{#2}}

% LaTeX commands to reduce the spacing above and below figures
% \renewcommand{\topfraction}{0.95}
% \setcounter{bottomnumber}{1}
% \renewcommand{\bottomfraction}{0.95}
% \setcounter{totalnumber}{3}
% \renewcommand{\textfraction}{0.05}
% \renewcommand{\floatpagefraction}{0.95}
% \setcounter{dbltopnumber}{2}
% \renewcommand{\dbltopfraction}{0.95}
% \renewcommand{\dblfloatpagefraction}{0.95}

%% usually not used, instead use the ordinary latex comment %
%\newcommand{\comment}[1]{} 
% \renewcommand{\paragraph}[1]{{\vspace{0.2em}\noindent\textbf{#1}}.}

% %% algorithm
% \makeatletter
% \newcommand{\StatexIndent}[1][3]{%
%   \setlength\@tempdima{\algorithmicindent}%
%   \Statex\hskip\dimexpr#1\@tempdima\relax}
% \makeatother

% \newdimen{\algindent}
% \setlength\algindent{1.5em}
% \algnewcommand\LeftComment[2]{%
% \hspace{#1\algindent}$\triangleright$ \eqparbox{COMMENT}{#2} \hfill %
% }

\renewcommand{\paragraph}[1]{{\vspace{2pt}\noindent\textit{#1}.}}

\begin{abstract}
As recent advances in large-scale Text-to-Image (T2I) diffusion models have yielded remarkable high-quality image generation, diverse downstream Image-to-Image (I2I) applications have emerged. Despite the impressive results achieved by these I2I models, their practical utility is hampered by their large model size and the computational burden of the iterative denoising process. In this paper, we propose a novel compression method tailored for diffusion-based I2I models. Based on the observations that the image conditions of I2I models already provide rich information on image structures, and that the time steps with a larger impact tend to be biased, we develop surprisingly simple yet effective approaches for reducing the model size and latency. We validate the effectiveness of our method on three representative I2I tasks: InstructPix2Pix for image editing, StableSR for image restoration, and ControlNet for image-conditional image generation. Our approach achieves satisfactory output quality with 39.2\%, 56.4\% and 39.2\% reduction in model footprint, as well as 81.4\%, 68.7\% and 31.1\% decrease in latency to InstructPix2Pix, StableSR and ControlNet, respectively.
%by applying it to InstructPix2Pix for image editing, StableSR for image restoration, and ControlNet for conditional image generation. Our approach achieves satisfactory output quality with 39.2\%, 56.4\% and 39.2\% reduction in model footprint, as well as 81.4\%, 68.7\% and 31.1\% decrease in latency to InstructPix2Pix, StableSR and ControlNet, respectively.
  %\keywords{Diffusion model compression \and Image-to-Image translation }
\end{abstract}

\setlength{\intextsep}{5pt}
\setlength{\columnsep}{5pt}

\section{Introduction}
\label{sec:intro}

In the advent of large-scale text-to-image (T2I) diffusion models such as DALL-E~\cite{dalle}, Stable Diffusion~\cite{ldm}, and Imagen~\cite{imagen}, there has been a dramatic improvement in image generation quality. This achievement has consequently opened up new opportunities across diverse applications, including image restoration~\cite{stablesr,diffbir}, image composition~\cite{tf-icon,crosscomposite,collagediffusion}, image editing~\cite{ip2p,pnp,dds,preditor,mdp}, conditional image synthesis~\cite{universal-guidance,mcm-diffusion,lawdiffusion,boxdiff,controlnet,t2i-adapter,freedom}, panorama generation~\cite{multidiffusion,diffcollage}, personalized generation~\cite{dreambooth}, creature generation~\cite{conceptlab}, and even 3D generation~\cite{dreamfusion,magic3d,fantasia3d,sjc}.
While these applications employing T2I models demonstrate unprecedented high-quality results, the extremely large parameter size combined with an iterative denoising process necessitates substantial computational resources, thus limiting their practicality. For instance, typical restoration networks generate images with fewer than 80 million parameters in a single feedforward pass~\cite{nafnet,uformer,gfpgan,hinet}.
Meanwhile, StableSR~\cite{stablesr}, which utilizes Stable Diffusion~\cite{ldm} for higher-quality image restoration, requires approximately 916 million parameters and at least 40 times longer latency, which is an unaffordable trade-off in many cases.
%Meanwhile, StableSR~\cite{stablesr}, which utilizes Stable Diffusion~\cite{ldm} for image restoration tasks, requires approximately 916 million parameters with dozens of iterations. While StableSR~\cite{stablesr} provides a higher quality output, it requires at least 40 times additional latency, which is an unaffordable trade-off in many cases. Saving memory footprints and reducing the computation cost of diffusion models are key factors for their widespread adoption.

Recently, numerous diffusion model compression approaches have been actively explored to reduce the memory and computational requirements of diffusion models.
% To reduce the memory and computational requirement of diffusion models, numerous diffusion model compression approaches have been actively explored recently.
%In response to this requirement, numerous studies have been actively conducted recently.
These approaches can be roughly categorized into two topics: reducing the number of denoising iterations~\cite{ddim,ddss,plms,dpm-solver,autodiff,vprediction,ondistillation} and reducing model footprint~\cite{bksdm,diffpruning}. 
% By focusing on the intrinsic features of diffusion models, these studies have proposed diverse task-agnostic optimization techniques.
By focusing on the innate characteristics of diffusion models, these studies have proposed diverse task-agnostic optimization techniques.
% However, their compression performance is still limited, as reducing the memory or computational budget beyond a certain threshold inevitably sacrifices the generative capabilities of diffusion models.
However, their compression performance remains insufficient for practical use in downstream tasks, and the potential for more effective compression methods applicable to I2I tasks has yet to be explored at all.

% In this work, we introduce a novel approach to reduce both memory footprint and latency of diffusion models for downstream Image-to-Image (I2I) applications.
% T2I diffusion models are designed to synthesize images including both their structures and details from random noise.
% In contrast, downstream I2I translation tasks such as image restoration, conditional image synthesis, and image editing use input images that provide substantial guidance on the structures of output images, so that the model needs to synthesize only details based on the input guidance.
% This offers significant potential for more aggressive compression of diffusion models beyond what existing task-agnostic methods offer, yet this potential remains unexplored so far.
% In this work, we explore such potential, and present a practical solution for I2I tasks that provides significant benefits in both latency and memory footprint over existing task-agnostic techniques while requiring a minimal cost.

In this work, we introduce a novel approach to reduce both memory footprint and latency of diffusion models for downstream Image-to-Image (I2I) applications.
While T2I diffusion models are designed to synthesize images including both their structures and details from random noise, downstream I2I translation tasks, such as image restoration, use input images that provide substantial guidance on the structures of output images.
This offers significant potential for more aggressive compression of diffusion models beyond what existing task-agnostic methods offer, yet this potential remains unexplored so far.
Therefore, we explore such potential, and present a practical solution for I2I tasks that provides significant benefits in both latency and memory footprint over existing task-agnostic techniques while requiring a minimal cost.

%In this work, we hypothesize that due to the relative simplicity of our target I2I tasks compared to pure generation tasks, there exist compression strategies uniquely suited to them that can provide significant benefits in both latency and memory footprint over existing task-agnostic techniques.

% Diffusion models are designed to synthesize images including both their structures and details from random noise.
% In contrast, downstream Image-to-Image (I2I) translation tasks such as image restoration, conditional image synthesis, and image editing use input images that provide substantial guidance on the structures of output images, so that the model needs to synthesize only details based on the input guidance.
% This offers significant potential for more aggressive compression of diffusion models beyond what existing task-agnostic methods offer, yet this potential remains unexplored so far.

% In this work, we introduce a novel approach to reduce both memory footprint and latency of diffusion models for downstream I2I applications.
% We hypothesize that due to the relative simplicity of our target I2I tasks compared to pure generation tasks, there exist compression strategies uniquely suited to them that can provide significant benefits in both latency and memory footprint over existing task-agnostic techniques.

Our approach comprises two surprisingly simple but effective components: depth-skip pruning and time-step optimization for reducing model size and latency, respectively.
Regarding depth-skip pruning, we first empirically verify that coarse layers of the denoising U-Net of a diffusion model, which primarily corresponds to coarse-grained features, contribute less to the output of downstream I2I operations.
Based on this, depth-skip pruning carefully prune less contributing coarse layers and fine-tune the model to effectively reduce the model size.

The time-step optimization method reduces the latency by searching for a reduced sequence of time steps of denoising iterations.
Specifically, the time-step optimization method searches for an optimal time-step sequence that produce high-quality outputs for a given number of time steps.
Unfortunately, finding an optimal time-step sequence is a challenging optimization problem as it involves integer variables, a nonlinear objective function, and a huge search space.
To overcome this challenge,
AutoDiffusion~\cite{autodiff} adopts the genetic algorithm, but it costs a huge amount of search time of a few days, and is prone to local minima.
Xue et al.~\cite{xue2024accelerating} mathematically derive a highly simplified approximation of the objective function, which can be optimized efficiently to find an approximate solution.
However, due to the simplicity of the approximation, their approach tends to find less optimal solutions.

For effective search for an optimal time-step sequence, our approach is based on the following intuition: depending on the I2I task, the distribution of the time steps with large impacts tend to be biased towards either the beginning or end of the iterative diffusion process as will be further discussed in \cref{ssec:m:timestep}.
Based on this intuition, we propose an extremely simple approach that aggressively reduces the search space to efficiently find an optimal time-step sequence.
Despite its simplicity, our experiments show that our approach achieves higher-quality results than previous approaches.

To reduce both memory footprint and latency, we apply depth-skip pruning and time-step optimization sequentially.
Our experiments show that the combination of the proposed depth-skip pruning and time-step optimization achieves satisfactory output quality with 60.8\% of parameters and 18.6\% of latency in InstructPix2Pix (IP2P)~\cite{ip2p}, 43.6\% of parameters and 31.3\% of latency in StableSR~\cite{stablesr}, and 60.8\% of parameters and 68.9\% of latency in ControlNet~\cite{controlnet} with canny-edge image as a condition input, respectively. 

\section{Related Work}
\label{sec:related works}
\subsubsection{I2I Downstream Tasks based on T2I Diffusion Models}
Thanks to the rich generative power of large-scale T2I models, transferring their generation capability to downstream I2I tasks have achieved state-of-the-art performance in various domains such as image inpainting~\cite{ldm}, depth-conditioned generation~\cite{ldm}, image restoration~\cite{stablesr,diffbir}, image editing~\cite{ip2p}, and conditional image synthesis~\cite{controlnet,t2i-adapter}.
%Since large-scale T2I models possess a rich generative prior for natural images, a straightforward way for transferring their generation capabilities to downstream I2I tasks have achieved state-of-the-art performance in various domains, such as text-conditioned image inpainting~\cite{ldm}, depth-conditioned generation~\cite{ldm}, image restoration~\cite{stablesr,diffbir}, image editing~\cite{ip2p}, and conditional image synthesis~\cite{controlnet,t2i-adapter}.
These downstream methods utilize the entire parameters and the complete denoising process, despite the relative simplicity of their tasks compared to the pure generation task that starts from Gaussian noise without any guidance images.
In this paper, we explore the compression potential of these I2I models, taking into account both model footprint and denoising iterations.

% \vspace{-3mm}
\subsubsection{Model Pruning of Diffusion Models}
Research on model pruning has primarily focused on pruning the architecture of monolithic end-to-end neural networks for image classification such as Convolutional Neural Networks (CNNs)~\cite{comp-cnn-1,comp-cnn-2,comp-cnn-3,comp-cnn-4} and Vision Transformer~\cite{comp-vit-1,comp-vit-2,comp-vit-3,comp-vit-4,comp-vit-5,comp-vit-6}.
Thus, applying these methods directly to diffusion models is challenging due to the intricate dynamics between the denoising network and time steps inherent in diffusion models.
%Research on model pruning has primarily focused on the network architecture, including pruning techniques for Convolutional Neural Networks (CNNs)~\cite{comp-cnn-1,comp-cnn-2,comp-cnn-3,comp-cnn-4} or Vision Transformer~\cite{comp-vit-1,comp-vit-2,comp-vit-3,comp-vit-4,comp-vit-5,comp-vit-6}. Applying these methods directly to diffusion models is challenging due to the intricate dynamics between the network and timesteps inherent in diffusion models.
Recently, a couple of works dedicated to diffusion models have been proposed~\cite{diffpruning,bksdm}.
Diff-Pruning~\cite{diffpruning} proposes a channel pruning method for diffusion models, which prunes a fixed amount of channels from each layer of the denoising network without considering the impacts of different layers, which can be different for different tasks.
BK-SDM~\cite{bksdm} analyzes the impact of different network blocks of Stable Diffusion~\cite{ldm}, and proposes three different versions of manually pruned models.
However, they do not propose an automatic pruning scheme that can be applied to other diffusion models with other metrics.
In contrast to these approaches, our depth-skip pruning is designed with a focus on downstream I2I tasks and offers a more principled approach to identifying redundant network blocks based on the quality constraints of a target task. 
%Diff-Pruning~\cite{diffpruning} proposes a channel pruning method for diffusion models, assuming a predefined channel sparsity for all network layers. Meanwhile, our depth-skip pruning aggressively prunes the entire coarse layers of the diffusion UNet that has minimal impact on the output for I2I downstream tasks.
%BK-SDM~\cite{bksdm} proposes a pruning method dedicated for large-scale T2I models using block-removal knowledge distillation. 
%While their method, similar to our depth-skip pruning, involves removing network components with less impact and subsequently retraining the pruned model, it offers only three fixed pruning specifications and the criteria for selecting network blocks to be removed appears somewhat arbitrary.
% Such restrictions give no room for flexible application in terms of network architectures, task-specific considerations, and quality-memory constraints that the user wants to set.
%In contrast, our depth-skip pruning is designed with a focus on downstream I2I models and offers a more principled approach to identifying the redundant network blocks, based on quality constraints of a target task. 

% \vspace{-3mm}
\subsubsection{Acceleration of Diffusion Models}
For fast sampling, alternative ODE or SDE samplers have been proposed \cite{ddim,dpm-solver,plms,analytic-dpm,zhang2022fast}, achieving a dramatic reduction in the number of iterations to fewer than a hundred. For further acceleration, parallel sampling methods~\cite{dsno,paradigm} have been proposed, but these methods cannot be directly applied to pretrained foundation models. 
OMS-DPM~\cite{oms-dpm} and T-stitch~\cite{t-stitch} present model scheduling methods that are applicable only when multiple pretrained diffusion models of varying sizes are available.
One notable branch is step distillation techniques~\cite{vprediction,ondistillation,add,instaflow,lcm,snapfusion}, which achieve feasible output quality with significantly small step numbers. Despite their effectiveness, they require a substantial amount of training time, approximately in the order of hundreds of V100 GPU days, to distill the pretrained knowledge of an original diffusion model to an accelerated model.

Another line of work is time scheduling approaches such as DDSS~\cite{ddss}, AutoDiffusion~\cite{autodiff} and Xue et al.~\cite{xue2024accelerating}.
These approaches aim to identify an optimal sequence of time steps within a predetermined number of denoising iterations. Such approaches offer a couple of benefits over the step distillation methods. They are computationally more efficient and simpler to implement, and crucially, preserve the full capabilities of the original model as they do not need to retrain a diffusion model, but allow to use the original model with a reduced time-step sequence.
This preservation significantly broadens its applicability to various downstream tasks unlike step distillation approaches.
Specifically, step distillation transforms the objective of a diffusion model from progressive denoising into tracking the mean posterior, which can impair the functionality of Classifier-Free Guidance (CFG) \cite{cfg} that is a vital control parameter in some applications such as IP2P \cite{ip2p}. It is also incompatible with certain downstream tasks that depend on the original denoising process of diffusion models \cite{stablesr,controlnet,t2i-adapter,diffbir}.
Our time-step optimization scheme also employs the time scheduling approach to support various downstream tasks.
%Therefore, for ensuring compatibility with a wide array of downstream applications, our timestep optimization technique also adopts a training-free way to speed up the denoising process, facilitating adaptable and efficient integration across different task variants.

\section{Methods}
\label{sec:methods}

In this section, we first briefly review the T2I diffusion model and transferred I2I model in \Sec{\ref{ssec:diffusionmodel}}.
% We then discuss the intuition and motivation behind our methods in \Sec{\ref{ssec:motivation}}.
We then introduce our depth-skip pruning for effectively reducing the model size in \Sec{\ref{ssec:m:depthskip}}, and our time-step optimization method to find an optimal time-step sequence in \Sec{\ref{ssec:m:timestep}}.
To reduce both model size and latency, depth-skip pruning and time-step optimization can be performed in any sequence, as the order has a negligible impact, as shown in the Supplemental Document (Tab.~S2). In our experiments, we initially perform depth-skip pruning, followed by time-step optimization.
%Lastly, we explain the way to combine these two methods in \Sec{\ref{ssec:m:combine}}.

% In this study, we introduce two distinct methods: depth-skip pruning and time-step optimization.
% Since the joint optimization of the depth-search and time-step search is highly challenging due to their enormous search space, we simply apply them sequentially, i.e., depth-skip pruning followed by time-step optimization.
% In the \cref{ssec:qual}, we will show that applying them sequentially consistently yields favorable outcomes. 

% \vspace{-2mm}
\subsection{Diffusion Models}
\label{ssec:diffusionmodel}

Diffusion models~\cite{ddpm,thermo,sde} are a class of generative models designed to convert Gaussian noise into a desired sample through iterative refinement using a denoising process guided by a neural network.
In T2I diffusion models, a noise prediction network $\epsilon_\theta$ is employed to estimate the noise component within the input image $x_t$ conditioned on time step $t$ and text prompt $\mathcal{P}$. For downstream I2I tasks utilizing T2I models, the model is retrained using a loss function:
\begin{equation}
  \begin{aligned}
    \mathcal{L} = ||\epsilon_{\theta}(x_t, c_I, \mathcal{P}, t) - \epsilon ||^2,
  \end{aligned}
\end{equation}
where $c_I$ represents an additional input image.
To accommodate the additional input, these approaches either fine-tune the diffusion model with minor modifications to the input network block~\cite{ldm,ip2p} or train a feature injection network while keeping the diffusion parameters fixed~\cite{stablesr,controlnet,t2i-adapter}.

\begin{figure*}[t]
  \centering
  \includegraphics[width=\linewidth]{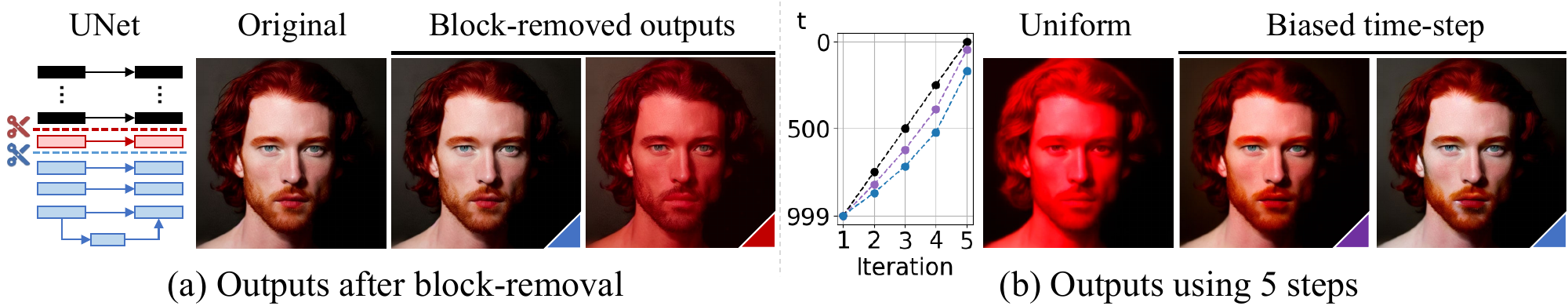}
  \vspace{-5mm}
  \caption{Motivations of our approach. (a) Even after removing the network layers beneath a certain depth, IP2P~\cite{ip2p}, a downstream I2I model, still produces a plausible result. (b) By focusing on earlier time steps, a feasible output can be obtained using only five denoising steps.}
  \label{fig:observations}
  \vspace{-0mm}
\end{figure*}

\begin{figure*}[t]
  \centering
  \includegraphics[width=\linewidth]{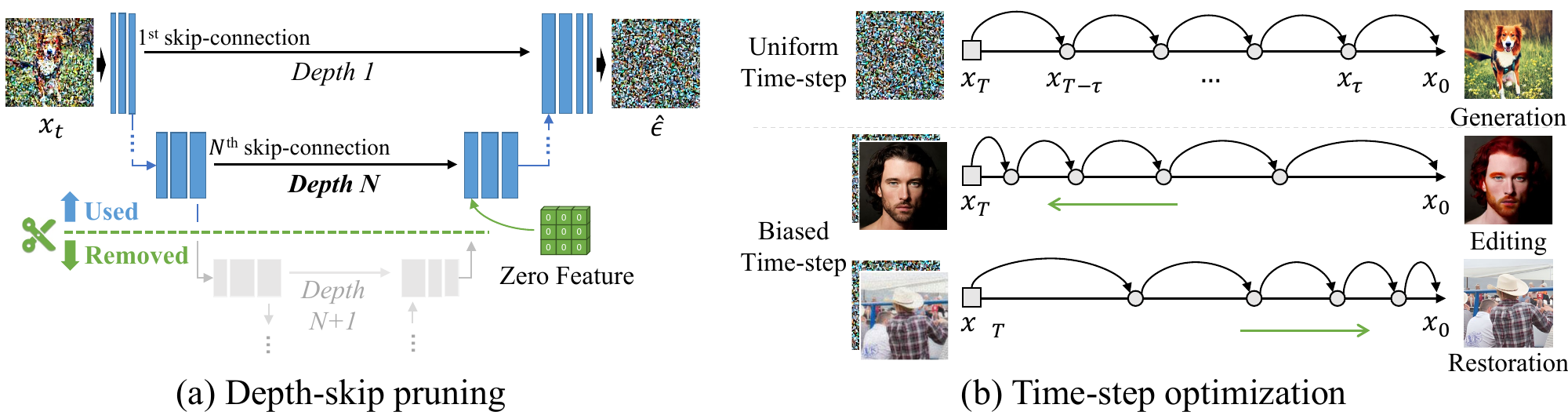}
  \caption{(a) Depth-skip pruning eliminates all layers deeper than a certain depth level, effectively reducing the model size. (b) Given a fixed number of time steps, our time-step optimization finds differently biased time step sequences for different I2I tasks. % Both approaches are highly effective in Image-to-Image (I2I) tasks that originate from a large-scale Text-to-Image (T2I) diffusion model.
  }
  \label{fig:main}
  \vspace{-3mm}
\end{figure*}

% \vspace{-2mm}
\subsection{Depth-Skip Pruning}
\label{ssec:m:depthskip}

%In this section, we present the intuition and details of the proposed depth-skip pruning method.
Our depth-skip pruning approach assumes that the denoising network of the target diffusion model is based on the U-Net~\cite{unet} architecture. 
In image generation tasks, the coarse layers of the U-Net are primarily responsible for creating the image structure~\cite{stylegan1,stylegan2,bigcolor}.
%When the U-Net~\cite{unet} is used to image generation tasks, the coarse layers of the network are primarily responsible for creating the image structure~\cite{stylegan1,stylegan2,bigcolor}. 
However, in I2I tasks, the image structure is already provided as input, such as a low-resolution image in image restoration task. As a result, we hypothesize that the deeper layers of the U-Net~\cite{unet} in the I2I model have a reduced impact on the final output.

To validate this assertion, we conduct an experiment in which we gradually remove the deepest network layers of IP2P~\cite{ip2p} and evaluate the output quality.
\cref{fig:observations}(a) presents the image editing results obtained by the IP2P models after removing the four and five deepest layers of the U-Net, which are indicated in blue and red, respectively.
Note that the models are not retrained after removing their layers.
% As the result shows, even if we simply remove deep layers beneath a certain depth, the output quality remains comparable to that of the original model, indicating that the deep layers have little impact on the output quality. This trend persists across different tasks.
As the result shows, even if we simply remove deep layers beneath a certain depth, the output quality remains comparable to that of the original model. This trend persists across different tasks, which indicates that the deep layers have little impact on the output quality.
Inspired by this, we introduce a depth-skip pruning which effectively reduces the model size.

\begin{wrapfigure}{R}{0.5\textwidth}
\small
\vspace{-0.0mm}
\begin{minipage}{0.5\textwidth}

\begin{algorithm}[H]
\caption{Depth-search}
\SetKw{KwFrom}{from}
\KwIn{
  input image $c_I$,
  prompt $\mathcal{P}$,
  maximum depth $d_{max}$,
  metric function $\mathcal{M}$
}
\KwOut{Optimal depth $d$}
  $d \leftarrow d_{max}$

  $x_T \sim \mathcal{N}(0,I)$ 

  \Repeat{$\operatorname{OverThreshold}(m)$}{

    $d \leftarrow d - 1$

    $x \leftarrow \operatorname{Sampler}_{DDIM}(x_T, c_I, \mathcal{P}; d)$ 

    $m \leftarrow \mathcal{M}(x)$
  }

  \label{alg:depthskip}
\end{algorithm}

\end{minipage}
\end{wrapfigure}

% \setlength{\intextsep}{5pt}%
% \setlength{\columnsep}{5pt}%
% \begin{wrapfigure}{R}{0.52\textwidth}
% \small
% \vspace{-1.0mm}
% \begin{minipage}{0.52\textwidth}

% \begin{algorithm}[H]
% \caption{Depth-search}
% \SetKw{KwFrom}{from}
% \KwIn{
%   input image $c_I$,
%   prompt $\mathcal{P}$,
%   maximum depth $d_{max}$,
%   metric function $\mathcal{M}$
% }
% \KwOut{Optimal depth $d$}
%   $d \leftarrow d_{max}$

%   $x_T \sim \mathcal{N}(0,I)$ 

%   \Repeat{$\operatorname{OverThreshold}(m)$}{

%     $d \leftarrow d - 1$

%     $x \leftarrow \operatorname{Sampler}_{DDIM}(x_T, c_I, \mathcal{P}; d)$ 

%     $m \leftarrow \mathcal{M}(x)$
%   }

%   \label{alg:depthskip}

% \end{algorithm}
% \end{minipage}
% \end{wrapfigure}

The key idea of the proposed method is to \emph{skip} certain network blocks located beyond a predetermined \emph{depth} of the skip-connection of the UNet.
For example, in a depth-8 model, denoted as $D$8, the network components beyond the 8th skip-connection are bypassed, as shown in \cref{fig:main}(a).
It is noteworthy that coarse-level network blocks typically have numerous channels, leading to substantial memory consumption. 
Depth-skip pruning removes these bulky blocks and allows us to effectively reduce the memory footprint while minimizing performance loss.

Our depth-skip pruning consists of two steps: depth-search and fine-tuning. 
In the depth-search step, we identify the target depth level for pruning by performing depth-skip from the deepest level upwards, using a predefined metric and threshold, until the quality threshold is met, as described in Alg.~\ref{alg:depthskip}.
Then, we fine-tune the pruned model to enhance its quality further.

\begin{table*}[t]
\centering

\caption{
Comparison between the single- and muti-depth search schemes on StableSR~\cite{stablesr}. 
The PSNR values are measured against the outputs obtained without depth-skip pruning.
%The percentages at ``Time'' and ``Param'', which indicates parameter, are proportion to requirements in full model.
%The PSNR is measured based on outputs without depth-skip.
%This results suggest that our single depth approach efficiently searches the near-optimal point compared to a multi-depth search strategy.
}

\vspace{-2mm}

\scalebox{0.70}{
\begin{tabular}{cccc|c|cc|ccc}
\Xhline{2\arrayrulewidth}
\multicolumn{4}{c|}{\multirow{2}{*}{Baseline (Single depth)}} & \multicolumn{1}{c|}{(a) Fix param. \& Min time}      & \multicolumn{2}{c|}{(b) Fix time \& Max quality}      & \multicolumn{2}{c}{(c) Fix quality \& Min time}     \\
\multicolumn{4}{c|}{}                           & \multicolumn{1}{c|}{( $\Delta$PSNR $<$ 0.2 dB )} & \multicolumn{2}{c|}{( $\Delta$Time $<$ 1\% )} & \multicolumn{2}{c}{( $\Delta$PSNR $<$ 0.2 dB )} \\ \hline
Depth    & PSNR     & Time(\%)    & Param(\%)   & $\Delta$Time(\%)         & $\Delta$PSNR       & $\Delta$Param(\%)    & $\Delta$Time(\%) & $\Delta$Param(\%) \\ \hline\hline
11       & 32.86    & 89.91       & 79.17       & -3.46                    & +0.02              & +9.47               & -3.46            & +0.00             \\
10       & 32.34    & 85.70       & 69.70       & -2.62                    & +0.34              & +9.47               & -4.59            & +9.47             \\
9        & 31.39    & 82.78       & 60.77       & -4.27                    & +1.08              & +18.40              & -5.72            & +8.93             \\
8        & 28.65    & 72.59       & 43.58       & -2.23                    & +1.53              & +26.13              & -6.29            & +26.13            \\ \Xhline{2\arrayrulewidth}
\end{tabular}

}

\label{tbl:ds-discuss}
\vspace{-3mm}
\end{table*}
\vspace{-5mm}
\subsubsection{Single- \textit{vs.} multi-depth search} 

Since we apply the same target depth level for all time steps (single-depth search), one might argue that our approach overly restricts the search space, potentially missing out on additional performance gains that could be achieved by using different depth levels for different time steps (multi-depth search). Here, we demonstrate that our single-depth search can find a near-optimal solution in a highly efficient manner compared to the multi-depth search.

To prove this, we compare the performances of our single-depth search scheme and the multi-depth search scheme.
For the multi-depth search, we find its optimal solutions using exhaustive search.
To mitigate the search overhead in this analysis, we confine the depth-skip levels from 7 to 12, and use 10 denoising iterations, and use the same depth levels for every two time steps, i.e., two consecutive time steps use the same depth level.

% Our method tightly confines the search space to a \emph{single} depth level for the entire time steps.
% Although this design decision minimizes search overhead,
% using different depth levels for different time steps may potentially yield additional performance gain.
% In the following, we demonstrate that this alternative method, referred as `multi-depth search', offers marginal gains despite its exponential growth of the search space.

% To this end, we compare the performances of our single-depth search scheme and the alternative multi-depth search scheme that finds a depth level for each time step.
% For the multi-depth search, we find its optimal solutions using exhaustive search.
% To mitigate the search overhead in this analysis, we confine the depth-skip levels from 7 to 12, and use 10 denoising iterations, and use the same depth levels for every two time steps, i.e., two consecutive time steps use the same depth level.

While the primary goal of depth-skip pruning is to reduce the model size,
it also affects the latency and output quality.
As a result, the multi-depth search may potentially find a solution that is more optimal in either quality or latency while having the same model size as the solution of the single depth-search approach.
Similarly, it may also find a superior solution in terms of model size or latency while having the same quality.
Thus, we analyze the performances in all the three aspects for a comprehensive analysis.

\paragraph{Optimizing quality or latency while fixing model size}
As diffusion models use a single denoising UNet for all time steps, the pruned model size is determined by the deepest depth across all time steps.
Having this in mind, to achieve the highest output quality while maintaining the same model size as the single-depth search, the solution of the multi-depth search must be the same as that of the single-depth search because maximizing the output quality necessitates utilizing as many network blocks as possible for all time steps.
On the other hand, we may find a solution with a smaller latency and the same model size using the multi-depth search if we accept a certain amount of quality degradation.
Nonetheless, the gain is small as shown in \cref{tbl:ds-discuss}(a). For the quality degradation of 0.2 dB, the gain in latency obtained by the multi-depth search is at most 4.27\%.

\paragraph{Optimizing model size or quality while fixing latency}
Another possibility is to use the multi-depth search to find a solution with a smaller model size or higher quality and the same latency as the result of the single-depth search.
However, finding a solution with a smaller model size while fixing the latency neither makes sensor nor is possible, as a smaller model inevitably leads to lower output quality and a smaller latency.
We may find a solution with a higher quality while fixing the latency using the multi-depth search.
However, in this case, the solution requires a significantly larger model size compared to that of our approach as shown in \cref{tbl:ds-discuss}(b).

\paragraph{Optimizing model size or latency while fixing quality}
We may use the multi-depth search to find a solution with a smaller model size and the same quality as the single-depth search.
However, again, this is impossible because a smaller model size inevitably causes lower output quality.
Finally, we may find a solution with a smaller latency and the same quality using the multi-depth search, but such a solution requires a substantially larger model size as shown in \cref{tbl:ds-discuss}(c).

This analysis indicates that our single-depth search scheme identifies solutions that are as effective as those found by multi-depth search, but with a considerably reduced search overhead.

\subsection{Time-step Optimization}
\label{ssec:m:timestep}

\begin{wrapfigure}{R}{0.6\textwidth}
\small
\vspace{-1.0mm}
\begin{minipage}{0.6\textwidth}

\begin{algorithm}[H]

\caption{Time-step optimization}
  \KwIn{
    step size $\eta$,
    metric function $\mathcal{M}$,
    signum function $\operatorname{sgn}$, 
    small value $\epsilon$,
    GT iteration $N$,
    target iteration $n$,
  }
  \KwOut{Optimal time-step $F_t(p_{prev}^s, n)$}

  $p \leftarrow 1$ ,
  $m \leftarrow \infty$ ,
  $x_T \sim \mathcal{N}(0,I)$ 

  $x_{uni} \leftarrow \operatorname{Sampler}_{DDIM}(x_T, c_I, \mathcal{P}, F_t(p,n))$ 

  $x_{pos} \leftarrow \operatorname{Sampler}_{DDIM}(x_T, c_I, \mathcal{P}, F_t(p + \epsilon,n) )$ 

  $x_{neg} \leftarrow \operatorname{Sampler}_{DDIM}(x_T, c_I, \mathcal{P}, F_t((p + \epsilon)^{-1},n))$ 

  $s \leftarrow \operatorname{sgn}(\mathcal{M}(x_{uni},x_{neg}) - \mathcal{M}(x_{uni}, x_{pos}))$

  $x^* \leftarrow \operatorname{Sampler}_{DDIM}(x_T, c_I, \mathcal{P}, F_t(p,N))$ 

  \Repeat{$m > m_{prev}$}{
    $m_{prev} \leftarrow m$ ,
    $p_{prev} \leftarrow p$ 

    $p \leftarrow p + \eta$ 

    $x \leftarrow \operatorname{Sampler}_{DDIM}(x_T, c_I, \mathcal{P}, F_t(p^s, n))$ 

    $m \leftarrow \mathcal{M}(x^*, x)$
  }
  
  \label{alg:skewedtimestep}
\end{algorithm}

\end{minipage}
\end{wrapfigure}

% \begin{algorithm}[t]

% \caption{Timestep optimization}
%   \KwIn{
%     step size $\eta$,
%     metric function $\mathcal{M}$,
%     signum function $\operatorname{sgn}$, 
%     small value $\epsilon$,
%     GT iteration $N$,
%     target iteration $n$,
%   }
%   \KwOut{Optimal timestep $F_t(p_{prev}^s, n)$}

%   $p \leftarrow 1$ ,
%   $m \leftarrow \infty$ ,
%   $x_T \sim \mathcal{N}(0,I)$ 

%   $x_{uni} \leftarrow \operatorname{Sampler}_{DDIM}(x_T, c_I, \mathcal{P}, F_t(p,n))$ 

%   $x_{pos} \leftarrow \operatorname{Sampler}_{DDIM}(x_T, c_I, \mathcal{P}, F_t(p + \epsilon,n) )$ 

%   $x_{neg} \leftarrow \operatorname{Sampler}_{DDIM}(x_T, c_I, \mathcal{P}, F_t((p + \epsilon)^{-1},n))$ 

%   $s \leftarrow \operatorname{sgn}(\mathcal{M}(x_{uni},x_{neg}) - \mathcal{M}(x_{uni}, x_{pos}))$

%   $x^* \leftarrow \operatorname{Sampler}_{DDIM}(x_T, c_I, \mathcal{P}, F_t(p,N))$ 

%   \Repeat{$m > m_{prev}$}{
%     $m_{prev} \leftarrow m$ ,
%     $p_{prev} \leftarrow p$ 

%     $p \leftarrow p + \eta$ 

%     $x \leftarrow \operatorname{Sampler}_{DDIM}(x_T, c_I, \mathcal{P}, F_t(p^s, n))$ 

%     $m \leftarrow \mathcal{M}(x^*, x)$
%   }
  
%   \label{alg:skewedtimestep}
% \end{algorithm}

Our time-step optimization scheme is inspired by the following observation:
earlier time steps are primarily involved in the generation of overall image structures incorporating the text prompt~\cite{ediffi}, while later time steps are mainly responsible for synthesizing image details~\cite{perception-step,analyze-step}.
Based on this observation, we hypothesize that later time steps are more influential in image restoration tasks like StableSR~\cite{stablesr}, whereas earlier time steps play a greater role in image editing like IP2P~\cite{ip2p}.
%
% In this section, we introduce the intuition and details of our time-step optimization method, which aims to accelerate the diffusion sampling process.
% To design our method, we first consider the properties of diffusion time steps: 
% earlier time steps are primarily involved in the generation of overall image structures incorporating the text prompt~\cite{ediffi}, while later time steps are mainly responsible for synthesizing image details~\cite{perception-step,analyze-step}.
% Based on these aspects, we hypothesize that later time steps are more influential in image restoration tasks like StableSR~\cite{stablesr}, whereas earlier time steps play a greater role in image editing like IP2P~\cite{ip2p}.
%
% We assume that the time-steps with a larger impact tend to be biased towards either the beginning or end of the iterative denoising process,
% based on the observations made in previous work~\cite{ediffi,perception-step,analyze-step},
% which suggest that early and later time steps have different roles in image synthesis.
% Specifically, earlier time steps are primarily involved in the generation of overall image structures incorporating the text prompt~\cite{ediffi}, while later time steps are mainly responsible for synthesizing image details~\cite{perception-step,analyze-step}.
% Therefore, we hypothesize that later time steps are more influential in image restoration tasks like StableSR~\cite{stablesr}, whereas earlier time steps play a greater role in image editing like IP2P~\cite{ip2p}.
%
To validate this, we conduct IP2P~\cite{ip2p} with only five steps at different intervals. In \cref{fig:observations}(b), the black dashed line represents IP2P~\cite{ip2p} generation using a uniform time sequence, while the purple and blue dashed lines represent non-uniform sequences with a focus on early time steps. As depicted in the figure, prioritizing earlier time steps yields viable results using only five steps. 
Although the impact of early and later time steps depends on tasks, we empirically observe similar phenomena in other I2I tasks. 
% This observation inspires our time-step optimization, where we aim to identify an optimal sequence of time steps given a fixed number of iterations.
Inspired by this observation, we design our time-step optimization method to find a biased sequence of time steps for each task, as illustrated in \cref{fig:main}(b).

Specifically, the time-step optimization aims to find an optimal sequence of time steps for a given number of time steps.
To design an effective and efficient parameterization for finding a biased sequence, we exploit the gamma curve formulation, which is defined as:
\begin{equation}
  \begin{aligned}
    F_t(\gamma,n) = T \cdot t^{\gamma},\quad \gamma > 0,
     \\ t=0, \frac{1}{n-1}, \frac{2}{n-1}, \cdots, 1
  \end{aligned}
\end{equation}
\noindent{}where $T$ represents the last time step, $\gamma$ is a parameter to control the shape of the gamma curve, and $n$ is the number of iterations.
If $\gamma > 1$, the generation process concentrates on the early time steps of generation, while if $0 < \gamma < 1$, it focuses on the later time steps. Then, our optimization problem becomes to find an optimal $\gamma$ that produces the output closest to the original sampling results using a small $n$.

% Specifically, the time-step optimization finds an optimal sequence of time-steps for a given number of time-steps.
% We design our time-step optimization to be highly efficient in terms of computational cost by aggressively limiting the search space to an one-dimensional space.
% As a result, our time-step optimization is as incredibly simple as an one-dimensional search.
% Nevertheless, as proven by our experiments in \cref{exp:ts}, it achieves higher-quality results than existing state-of-the-art methods such as AutoDiffusion~\cite{autodiff}, while being at least 62 times faster.

% Our time-step optimization is based on the intuition presented in \cref{ssec:motivation}.
% As illustrated in \cref{fig:main}(b), our time-step optimization finds a biased sequence of time steps for each task.
% Specifically, we search for an optimal time-step sequence in the one-dimensional search space of slopes $\gamma$ of gamma curves where a gamma curve is defined as:
% \begin{equation}
%   \begin{aligned}
%     F_t(\gamma,n) = T \cdot t^{\gamma},\quad \gamma > 0, \quad
%      t=0, \frac{1}{n-1}, \frac{2}{n-1}, \cdots, 1
%   \end{aligned}
% \end{equation}
% \noindent{}where $T$ represents the last time step, $\gamma$ is a parameter of the gamma curve, and $n$ is the number of iterations.
% If $\gamma > 1$, the generation process concentrates on the early time steps of generation, while if $0 < \gamma < 1$, it focuses on the later time steps. Then, our optimization problem becomes to find an optimal $\gamma$ that produces the output closest to the original sampling results using a small $n$.

While this simple strategy already provides comparable or outperforming results to previous state-of-the-art approaches \cite{autodiff,xue2024accelerating}, it is still limited due to the fixed nature of the first and last time steps.
%However, this straightforward optimization often leads to unsatisfactory results due to the fixed nature of the first and last time step. 
To further improve the performance, we introduce a scale-down mechanism for the gamma curve.
%To address this limitation, we apply a scale-down mechanism for the gamma curve. 
Specifically, we scale down the gamma curve toward $T$ when $\gamma < 1$ proportional to decrease of $\gamma$ value, and vice versa.
The formal definition is as follows:

\begin{equation}
  \begin{aligned}
    F_t(\gamma,n) = T \cdot t'^{\gamma}, \quad t' = \frac{T\cdot t-t_l}{t_u - t_l}\\
    (t_l,\space t_u)= 
    \begin{cases}
        (0,\space T+\alpha(\gamma-1)),&  \gamma \geq 1\\
        (\alpha(1 - 1/\gamma),\space T),& \gamma < 1
    \end{cases}
  \end{aligned}
\end{equation}
where $\alpha$ is a coefficient for scale strength. 
This adjustment allows for greater flexibility and potentially more effective optimization of the time steps. 

The search process consists of two stages.
Firstly, we determine whether $\gamma$ increases or decreases by evaluating which direction yields better outputs. 
Then, we perform a greedy search by progressively increasing or decreasing the value of $\gamma$ until no further improvement in quality is observed, as described in \Alg{\ref{alg:skewedtimestep}}.
Our time-step optimization is designed to be highly efficient in terms of computational cost by aggressively limiting the search space to one dimension. Despite this simplification, as demonstrated in our experiments (\cref{exp:ts}), it achieves higher-quality results than existing state-of-the-art methods like AutoDiffusion~\cite{autodiff}, while being at least 62 times faster.

% \subsection{Combination of depth-skip pruning and time-step optimization}
% \label{ssec:m:combine}

% In this study, we introduce two distinct methods: depth-skip pruning and time-step optimization.
% Since the joint optimization of the depth-search and time-step search is highly challenging due to their enormous search space, we simply apply them sequentially, i.e., depth-skip pruning followed by time-step optimization.
% In the \cref{ssec:qual}, we will show that applying them sequentially consistently yields favorable outcomes. 

% It is important to highlight that, in this study, we introduce two distinct methods: depth-skip pruning and time-step optimization instead of a single unified method. Searching for an optimal solution in both memory footprint and latency is highly challenging due to their enormous search space. Instead, in this paper, we aim at developing practical solutions that can be easily applied with minimal cost. Our experiments show that applying them sequentially consistently yields favorable outcomes. In the following subsections, we will provide detailed explanations of each method.

\begin{figure}[t!]
  \centering
  \includegraphics[width=0.95\linewidth]{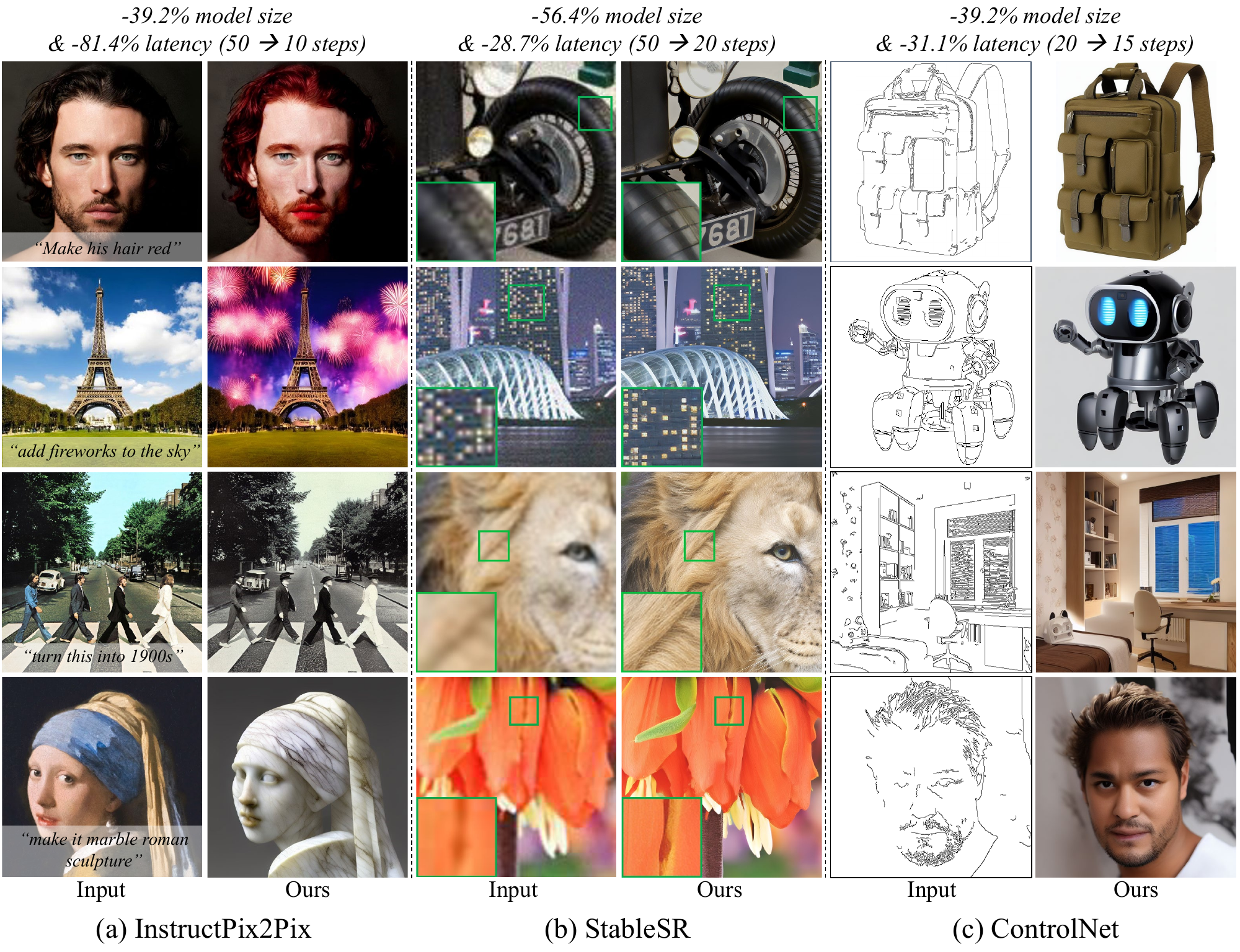}
  \vspace{-2mm}
  \caption{
  Qualitative examples of our depth-skip pruning and time-step optimization on IP2P~\cite{ip2p}, StableSR~\cite{stablesr}, and ControlNet~\cite{controlnet}.
  %We achieve satisfactory outcomes with significantly reduced model size and latency, as described on top of the caption. The baseline step numbers are defined as 50 steps in IP2P~\cite{ip2p} and StableSR~\cite{stablesr}, and 20 steps in ControlNet~\cite{controlnet}, as originally specified in their respective applications.
  }
  \vspace{-4mm}
  \label{fig:qual}
\end{figure}

\section{Experiments}
\label{sec:experiments}

In this section, we validate the effectiveness of our compression method by applying it to IP2P~\cite{ip2p} for image editing, StableSR~\cite{stablesr} for image restoration, and ControlNet~\cite{controlnet} for image-conditioned image generation. In the case of ControlNet, we use canny edge maps as input condition in our experiments.
The baseline step numbers used are the same as originally specified in their respective applications: 50 steps for IP2P~\cite{ip2p} and StableSR~\cite{stablesr}, and 20 steps for ControlNet~\cite{controlnet}.
When discussing the model size and latency, we only consider those of the diffusion U-Net, excluding the text-encoder~\cite{clip}, auto-encoder~\cite{vqgan}, and additional adapter networks~\cite{controlnet} unless specified otherwise.
We refer the readers to Sec.~S1 in the Supplemental Document for more details.

\subsection{Qualitative and Quantitative Comparisons}
\label{ssec:qual}
\Fig{\ref{fig:qual}} shows results of the models of different tasks compressed by the proposed depth-skip pruning and time-step optimization.
Compared to the original models of IP2P~\cite{ip2p}, StableSR~\cite{stablesr}, and ControlNet~\cite{controlnet},
our compressed models use only 60.8\%, 43.6\%, and 60.8\% of the parameters, respectively, and their latencies are reduced to 18.6\%, 31.3\%, and 68.9\%, respectively.
\cref{tbl:timeconsum} reports a quantitative comparison of the latencies and model sizes of the original models and their compressed results.
In this comparison, we also report the total sizes and latencies including those of the VAE, text encoder and adapter networks.
As the results show, despite the much smaller model sizes and latencies, the compressed models successfully produce visually pleasing results, clearly indicating that our method effectively reduces both model size and latency while preserving the original generative power required for each task.

\begin{table}[t]
\centering
\caption{
Latency and the number of parameters including VAE, text encoder, and adapter network. ``U-Net+'' indicates the union of the U-Net and additional adapter network. The compression includes both depth-skip pruning and time-step optimization. The unit of latency is seconds.
}
\vspace{-2mm}

\scalebox{0.75}{

\begin{tabular}{c|ccc|ccc}
\Xhline{2\arrayrulewidth}
% \multicolumn{1}{l|}{} & \multicolumn{3}{c|}{Latency (U-Net+/Total/Iter.)}       & \multicolumn{3}{c}{Parameter (U-Net+/Total)}     \\
\multicolumn{1}{l|}{} & \multicolumn{3}{c|}{Latency}       & \multicolumn{3}{c}{Parameter}     \\
\multicolumn{1}{l|}{} & \multicolumn{3}{c|}{(U-Net+/Total/Iteration)}       & \multicolumn{3}{c}{(U-Net+/Total)}     \\ 
\multicolumn{1}{l|}{}        &  Original  & \phantom{s} Compressed \phantom{s}  & Total Reduction & Original & \phantom{s} Compressed \phantom{s} & Total Reduction \\\hline\hline
IP2P~\cite{ip2p}             & 6.31/6.54/50 & 1.17/1.40/10 & 78.6\%    & 859M/1066M  & 522M/729M  & 31.6\%    \\
StableSR~\cite{stablesr}     & 2.81/2.94/50 & 0.88/1.01/20 & 65.7\%    & 969M/1176M  & 452M/658M  & 44.0\%    \\
ControlNet~\cite{controlnet} & 1.57/1.75/20 & 1.08/1.26/15 & 28.0\%    & 1220M/1427M & 883M/1090M & 23.6\%   \\
\Xhline{2\arrayrulewidth}
\end{tabular}
\label{tbl:timeconsum}
}

\end{table}

\begin{figure}[t!]
  \centering
  \includegraphics[width=0.95\linewidth]{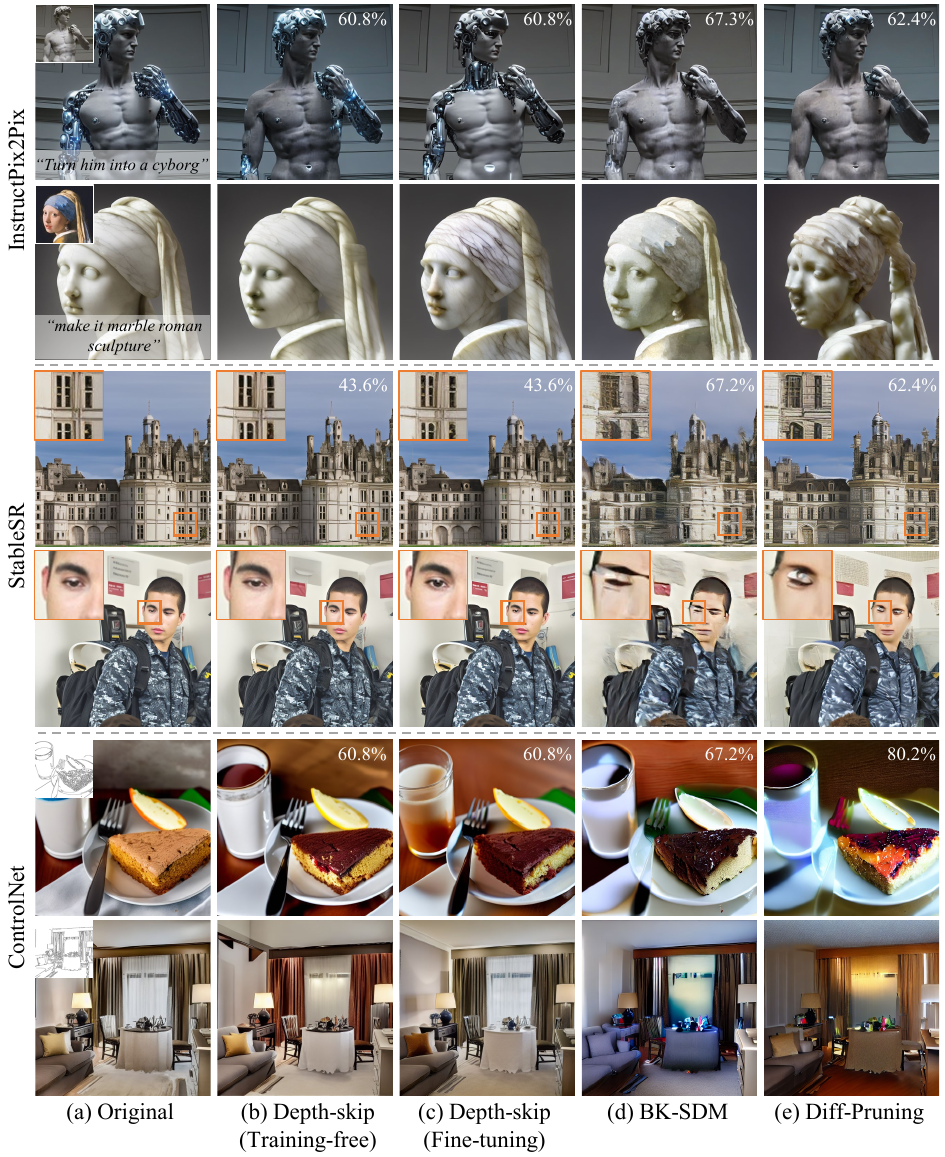}
  \vspace{-2mm}
  \caption{Comparison of the depth-skip pruning and previous pruning methods. The number on the top-right side in each image denotes the pruned model size.
  %Our results show more visually pleasing results, even with a smaller model size and without fine-tuning.
  }
  \label{fig:ds-qual}
  \vspace{-5mm}
\end{figure}

% \vspace{-2mm}
\subsection{Evaluation of Depth-skip Pruning}
\label{exp:ds}

We compare the proposed depth-skip pruning with state-of-the-art pruning methods tailored for diffusion models: Diff-pruning~\cite{diffpruning} and BK-SDM~\cite{bksdm}.
For fine-tuning, we followed the original training strategy except for ControlNet~\cite{controlnet}, as the LAION~\cite{laion} dataset used for ControlNet is no longer publicly available. Instead, we used the COCO~\cite{coco} dataset.
The optimal depths searched by our depth-search algorithm are $D9$ (60.8\% of parameters) in IP2P~\cite{ip2p} and ControlNet~\cite{controlnet}, and $D8$ (43.6\% of parameters) in StableSR~\cite{stablesr}, respectively.
We refer to Sec. S2 in the Supplemental Document for experimental results involving extreme depth-skip pruning, such as $D6$ (15.6\% of parameters) models.

\begin{table}[t!]
    \centering

    \caption{
    Quantitative comparisons of the depth-skip pruning and other pruning methods for StableSR~\cite{stablesr} and ControlNet~\cite{controlnet}. % Note that our depth-skip pruning outperforms previous pruning methods even without fine-tuning.
    }
    \vspace{-1mm}
    \begin{subtable}[c]{1.0\textwidth}
    \centering
    \scalebox{0.8}{
        \begin{tabular}{lccccc}
        \Xhline{2\arrayrulewidth}
        Model                                       & Steps   & FID$\downarrow$  & PSNR$\uparrow$   & LPIPS$\downarrow$  & Parameter           \\ \hline\hline
        Diff-pruning~\cite{diffpruning}             & 50      & 38.70            & 21.20            & 0.483              & 578M\space(62.4\%)  \\
        BK-SDM~\cite{bksdm}                         & 50      & 64.46            & 21.45            & 0.492              & 615M\space(67.2\%)  \\
        Depth-skip ($D$9) without fine-tuning       & 50      & {\bf 28.55}      & 21.54            & {\bf 0.441}        & 557M\space(60.8\%)  \\ 
        Depth-skip ($D$8) without fine-tuning       & 50      & 40.25            & 21.40            & 0.467              & 400M\space(43.6\%)  \\ 
        Depth-skip ($D$8)                           & 50      & 30.15            & 21.48            & 0.449              & 400M\space(43.6\%)  \\ 
        Depth-skip ($D$8) + Time-step optimization  & 20      & 32.31            & {\bf 21.82}      & 0.457              & 400M\space(43.6\%)  \\ \hline
        Original                                    & 50      & 27.70            & 21.51            & 0.437              & 917M\space(100\%)   \\ 
        \Xhline{2\arrayrulewidth}
        \end{tabular}
    }
    \caption{StableSR~\cite{stablesr}}
    \end{subtable}
    
    \begin{subtable}[c]{1.0\textwidth}
    \centering
    \scalebox{0.75}{
        \begin{tabular}{lccccc}
        \Xhline{2\arrayrulewidth}
        Model                                        & Steps & FID $\downarrow$  & CLIP-Score$\uparrow$ & CLIP-a$\uparrow$ & Parameter \\ \hline\hline
        Diff-pruning~\cite{diffpruning}              & 20    & 28.52             & 28.91          & 5.01             & 687M\space(80.2\%) \\
        BK-SDM~\cite{bksdm}                          & 20    & 29.66             & 29.08          & 4.87             & 576M\space(67.2\%) \\
        Depth-skip ($D$9) without fine-tuning        & 20    & 21.18             & 30.29          & {\bf 5.93}       & 521M\space(60.8\%) \\ 
        Depth-skip ($D$9)                            & 20    & {\bf 17.64}       & {\bf 30.46}    & 5.88             & 521M\space(60.8\%) \\
        Depth-skip ($D$9) + Time-step optimization   & 15    & 18.65             & 30.37          & 5.87             & 521M\space(60.8\%) \\ \hline 
        Original fine-tuned on the COCO dataset      & 20    & 17.52             & 30.57          & 5.91             & 857M\space(100\%)  \\
        Original                                     & 20    & 19.88             & 30.42          & 6.10             & 857M\space(100\%)  \\ 
        \Xhline{2\arrayrulewidth}        
        \end{tabular}
        }
    \caption{ControlNet~\cite{controlnet}}
    \end{subtable}
    
    \vspace{-7mm}
    \label{tbl:quant}
\end{table}

\begin{figure}[t!]
     \centering
     \begin{subfigure}[b]{0.495\linewidth}
         \centering
         \includegraphics[width=1.0\linewidth]{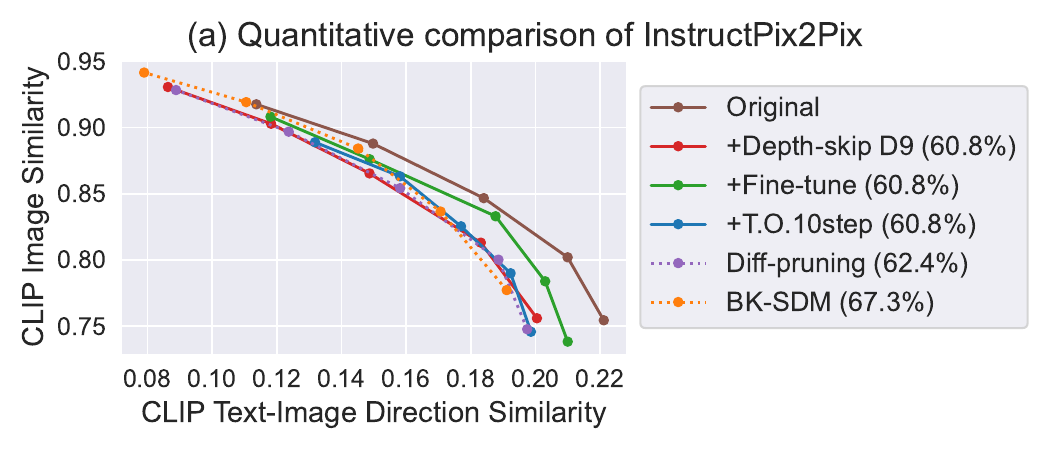}
         % \caption{Various configurations of IP2P~\cite{ip2p}}
         \label{fig:quant-fig-a}
     \end{subfigure}
     \hfill
     \begin{subfigure}[b]{0.495\linewidth}
         \centering
         \includegraphics[width=1.0\linewidth]{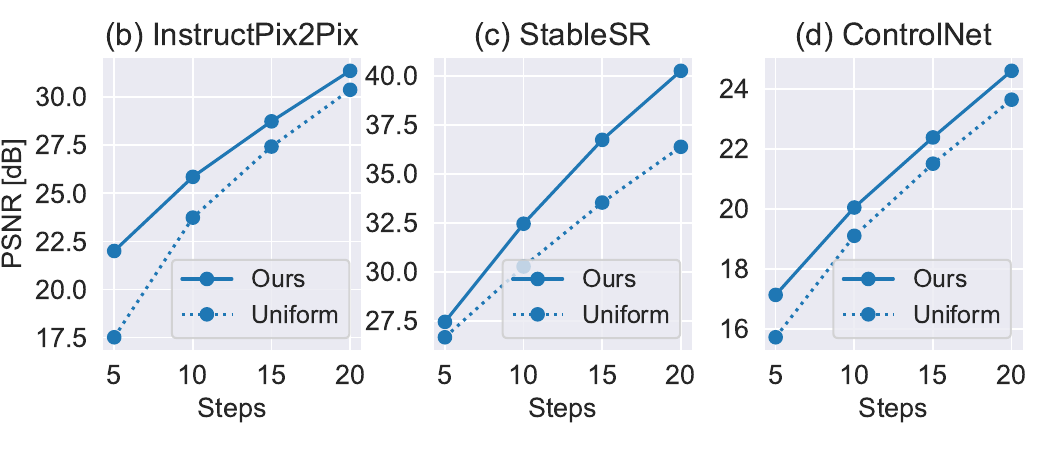}
         % \caption{Quantitative comparisons on acceleration}
         \label{fig:quant-fig-b}
     \end{subfigure}
    \vspace{-10mm}
        \caption{
        (a) Quantitative comparison of the depth-skip pruning and the other methods on IP2P~\cite{ip2p}.
        ``T.O.'' denotes the time-step optimization.
        (b-d) Comparison between the time-step optimization and uniform sampling.
        %We shows trade-off between consistencies of input image and editing for different pruning and time-step configurations applied to IP2P~\cite{ip2p}. ``T.O.'' denotes the use of our time-step optimization. (b-d) We plots step numbers and corresponding PSNR scores with original sample for uniform time-step and our time-step optimization.
        }
        \label{fig:quant-fig}
        \vspace{-4mm}
\end{figure}

\paragraph{Qualitative comparison}
\Fig{\ref{fig:ds-qual}} shows a qualitative comparison of depth-skip pruning with and without fine-tuning, and the previous pruning methods.
% The previous methods involve fine-tuning after pruning.
\Fig{\ref{fig:ds-qual}}(a) shows the results of the original models for each task without any pruning.
Compared to the original results, the previous methods often produce semantically incorrect results with artifacts despite their larger model sizes than ours.
On the other hand, our depth-skip pruning shows more visually pleasing results compared to the previous methods, even with a smaller model size and without fine-tuning.
Note that the results of our depth-skip pruning are not exactly the same as the original results due to the pruned layers and fine-tuning.
Nevertheless, our methods produce semantically correct results, showing that the generative capabilities of the original models are well preserved.
% Notably, our depth-skip pruning significantly outperforms the previous methods, \emph{even with a smaller model size and without fine-tuning}.

\paragraph{Quantitative comparison}
For quantitative comparison, we first compare the performances of the previous approaches and ours on StableSR~\cite{stablesr}.
To this end, we use the super-resolution validataion dataset of StableSR, which is generated from the DIV2K dataset~\cite{div2k}.
\cref{tbl:quant}(a) shows the quantitative comparison.
We measure FID~\cite{fid}, PSNR and LPIPS~\cite{lpips} scores for evaluation.
% For the PSNR and LPIPS~\cite{lpips} scores, we measure them against the results from the original StableSR model.
As the table shows, $D$9 without fine-tuning and $D$8, both of which are our depth-skip pruning results, achieve the best FID~\cite{fid}, PSNR and LPIPS~\cite{lpips} scores close to the scores of the original model, significantly outperforming Diff-pruning~\cite{diffpruning} and BK-SDM~\cite{bksdm} even though they have smaller model sizes and $D$9 does not use fine-tuning.
The table also shows the effect of the fine-tuning step in the depth-skip pruning. By comparing $D$8 before and after fine-tuning, it is evident that fine-tuning significantly enhances performance, enabling $D$8 to match the original model's performance with less than half of the original model's size.
Finally, although time-step optimization results in a minor quality degradation due to the reduction of iterations by more than half, it still surpasses the performance of the previous methods.

We also conduct a quantitative evaluation on ControlNet~\cite{controlnet}.
For evaluation, we use the COCO validation set~\cite{coco} for ControlNet.
For the input text prompts required for the ControlNet models, we generate text prompts using BLIP~\cite{blip}.
\cref{tbl:quant}(b) shows the quantitative comparison.
As mentioned earlier, we use the COCO dataset for fine-tuning instead of the LAION dataset~\cite{laion}. Thus, we also compare the result of the original model fine-tuned on the COCO dataset.
Similar to the results in \cref{tbl:quant}(b), despite their smaller model sizes, our results achieve the best scores in all quality metrics regardless of fine-tuning and time-step optimization.

Finally, we conduct a quantitative comparison on IP2P~\cite{ip2p}.
For evaluation, we follow the protocol of IP2P~\cite{ip2p}.
Specifically, we measure the CLIP image similarity scores~\cite{clip} and CLIP text-image direction similarity scores~\cite{stylegannada}. As IP2P allows the control of the editing strength using the CFG~\cite{cfg} parameter, we plot the scores for different image CFG parameter values ranging from 1.0 to 1.8.
\cref{fig:quant-fig}(a) shows the quantitative comparison.
The solid lines in this figure display the quantitative results where the depth-skip, fine-tuning and time-step optimization with 10 steps are successively applied, and the dotted lines show the results of previous pruning method. 
Our depth-skip pruning without fine-tuning shows comparable results to other pruning methods.
After fine-tuning, which recovers the quality degradation of model pruning, our pruning method outperforms the other methods by a large margin.

\subsection{Evaluation of Time-step Optimization}
\label{exp:ts}
% \Fig{\ref{fig:time-qual}} shows qualitative comparisons for our optimized timesteps and uniform timesteps, demonstrating the effectiveness of our approach. Specifically, ours displays similar results only using half the steps in IP2P~\cite{ip2p}, and shows outcomes closer to the original output with smaller number of steps in ControlNet~\cite{controlnet}.
We evaluate the performance of the proposed time-step optimization.
For evaluation of the proposed method, we apply time-step optimization to the original models without applying depth-skip pruning. Also, we randomly sample 100 images from the training dataset for the search process, and employ a bias coefficient of $\alpha = 30$.

\begin{figure}[t]
  \centering
  \includegraphics[width=0.95\linewidth]{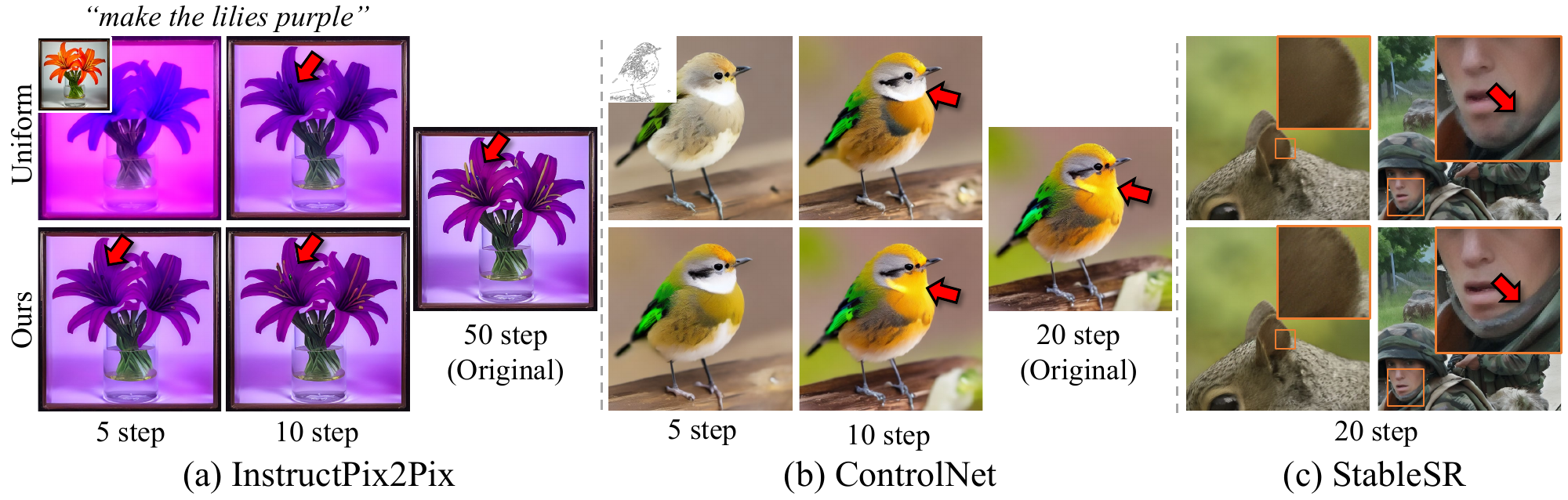}
  \vspace{-1mm}
  \caption{
Comparison between the time-step optimization and uniform sampling.
%(a,b) Ours shows the outcome closer to the original output than those with uniform time-step, and (c) produces better texture and edge elements.
}
  \label{fig:time-qual}
  \vspace{-0mm}
\end{figure}

\cref{fig:time-qual} shows a qualitative comparison between the results of our time-step optimization and the uniform sampling strategy that samples evenly distributed time steps.
For all the tasks, our approach produces superior results than the uniform sampling strategy.
Specifically, in the case of IP2P~\cite{ip2p} and ControlNet~\cite{controlnet}, our method produces results that are similar to the results of the original models even with only five steps.
On the other hand, the outputs of the uniform sampling scheme quickly degrade as the number of time steps decreases.
In the case of StableSR~\cite{stablesr}, our results show accurately restored high-frequency details, while the uniform sampling fails to restore such details.

\cref{fig:quant-fig}(b-d) visualizes PSNR values of the time-step optimization and uniform sampling strategies with respect to different numbers of iterations.
The PSNR values of the outputs from the optimized time steps are measured against the results of the original models with 50 iterations using DDIM~\cite{ddim} deterministic process. 
Also, we measure the metric based on a random selection of 1,000 images from the validation dataset for each task.
% The PSNR values are measured using the dataset specific to each task, as previously described.
Across all the iteration numbers, our time-step optimization consistently yields higher PSNRs for all the tasks.

\cref{tbl:autodiff} compares the output qualities and search times of our time-step optimization with those of previous state-of-the-art time scheduling approaches for diffusion models: AutoDiffusion~\cite{autodiff} and Xue et al.'s method~\cite{xue2024accelerating}.
As Xue et al.'s method is based on a highly simplified mathematical approximation for efficient time scheduling, it takes only a few seconds for five time steps, and less than a minute for 10 time steps. Nevertheless, due to its approximation, its output quality is the lowest among the compared methods.
On the contrary, AutoDiffusion takes tens of hours to search for optimal time steps due to its reliance on the genetic algorithm. Despite the lengthy search duration, it still lags behind our method in output quality, as the genetic algorithm tends to get trapped in local minima. In contrast, thanks to its constrained yet effective search space, our method only requires 10 to 40 minutes and consistently delivers superior output quality for all the cases.
More analyses and details can be found in Sec.~S3 and S4 in the Supplemental Document.

% presents a plot of the speeds required to achieve comparable PSNR with the original sample for both uniform and our optimized time-step. Our method outperforms the uniform time-step strategy across various levels of accuracy. Specifically, our optimal time steps for IP2P~\cite{ip2p} using 5 steps is 1.6 times faster than the uniform approach.
% % Additionally, the figure suggests that given the high-level PSNR achieved with a relatively small iteration in a uniform way, the conventional practice of using 50 iterations is unnecessarily excessive for these I2I tasks.
% We also observed that the optimized time-step has little change after applying depth-skip compression. On the other hand, it varies in changes of CFG~\cite{cfg} values. Please refer to the Supplementary Document for more detailed analyses.

\begin{table}[t]
\centering
\caption{
Quantitative comparison of time-step optimization with previous methods.
%The scores represent the PSNR metrics measured using 50 iterations.
}
\vspace{-2mm}

\scalebox{0.75}{

\begin{tabular}{c|c|ccc|ccc|ccc}
\Xhline{2\arrayrulewidth}
& \multicolumn{1}{l|}{} & \multicolumn{3}{c|}{InstructPix2Pix~\cite{ip2p}}          & \multicolumn{3}{c|}{StableSR~\cite{stablesr}}                 & \multicolumn{3}{c}{ControlNet~\cite{controlnet}}               \\
& \# Steps                 & Ours           & AutoDiff. & Xue et al. & Ours           & AutoDiff. & Xue et al. & Ours           & AutoDiff. & Xue et al. \\ \hline\hline
PSNR & 5                    & \textbf{22.00} & 20.64         & 14.35      & \textbf{27.46} & 26.58         & 25.83      & \textbf{17.14} & 16.83         & 13.89      \\
(dB) & 10                   & \textbf{25.86} & 24.79         & 20.62      & \textbf{32.46} & 29.03         & 27.71      & \textbf{20.05} & 19.23         & 16.23      \\
 \hline
Search & 5                    & 38.7m & 40.5h         & 3.2s      & 9.5m & 16.3h         & 3.2s      & 14.1m & 31.2h         & 3.2s      \\
time & 10                   & 30.9m & 75.1h         & 53.1s      & 11.1m & 27.1h         & 53.1s      & 18.0m & 65.1h         & 53.1s      \\ 
\Xhline{2\arrayrulewidth}
\end{tabular}

}
\vspace{-3mm}
\label{tbl:autodiff}
\end{table}

%\input{tbls/autodiff-time+time-quant}

% We also compare our time-step optimization with AutoDiffusion~\cite{autodiff} and Xue et al.~\cite{bksdm}'s method that aim to find optimal time sequences to accelerate the denoising process. \Tbl{\ref{tbl:autodiff}} illustrates that our method outperforms the previous time-step optimization methods. In terms of optimization time, our method takes less than an hour, whereas AutoDiffusion~\cite{autodiff} generally requires more than a day, as indicated in \Tbl{\ref{tbl:autodiff-time}}.
% Although Xue et al.'s method has an extremely fast search time, it shows poor quality, as indicated in \Tbl{\ref{tbl:autodiff}}.
% For more detailed analysis and comparisons, please refer to the Supplementary Document.

% \input{figs/autodiff-layer.tex}
% \vspace{-0mm}
% \paragraph{Layer-skip of AutoDiffusion}
% AutoDiffusion~\cite{autodiff} optimizes both the time steps and skip configuration of network layers similar to our compression method. However, their layer-skip search does not reduce network parameters, as demonstrated in \Fig{\ref{fig:layer-skip}}. Also, the acceleration achieved by the layer-skip is marginal, and there is a noticeable performance trade-off, as shown in \Tbl{\ref{tbl:layer-skip}}.
%%% CONCLUSIONS %%%%
\section{Conclusion} % (fold)
\label{sec:conclusion}
In this paper, we introduced a novel compression method for downstream I2I diffusion models,
which consists of depth-skip pruning and time-step optimization for reducing the memory footprint and latency, respectively.
Despite their simplicity, our experiments show that they significantly outperform previous state-of-the-art task-agnostic pruning and time scheduling approaches.
%our method with IP2P~\cite{ip2p}, StableSR~\cite{stablesr} and ControlNet~\cite{controlnet}, achieving practical results, including 39.2\%, 56.4\% and 39.2\% reduction in model footprint, as well as 81.4\%, 68.7\% and 31.1\% decrease in latency, respectively.

%In this paper, we introduce a novel compression method for downstream I2I models that combines depth-skip pruning, reducing model size by pruning coarse network layers, and time-step optimization, finding the optimal time-step sequence using a scaled-gamma curve. We validate our method with IP2P~\cite{ip2p}, StableSR~\cite{stablesr} and ControlNet~\cite{controlnet}, achieving practical results, including 39.2\%, 56.4\% and 39.2\% reduction in model footprint, as well as 81.4\%, 68.7\% and 31.1\% decrease in latency, respectively.

\vspace{-0mm}
\paragraph{Limitation \& Future work}
Our depth-skip pruning assumes that the denoising network has a U-Net~\cite{unet}-based architecture. Therefore, the pruning method would be unavailable to other diffusion models with different network architectures, such as transformers~\cite{dit}. Developing a compression method applicable to various network architectures could be a promising future direction.

% A fundamental limitation of timestep optimization~\cite{ddss,autodiff,xue2024accelerating} is the relatively modest acceleration it offers compared to the step distillation approach. Therefore, our future work will concentrate on developing a distillation-based acceleration method that can comprehensively accommodate a wide range of downstream task model variants.

% While timestep optimization demonstrates effectiveness in small iterations, it faces challenges in achieving comparable performance with significantly reduced iterations, which is a fundamental limitation of optimizing the time schedule~\cite{ddss,autodiff}. Therefore, our future work will focus on achieving comparable quality with significantly fewer iterations by applying a step distillation method that is broadly applicable to various downstream models.

\paragraph{\small {\bf Acknowledgements}}
This work was supported by Institute of Information \& communications Technology Planning \& Evaluation (IITP) grants funded by the Korea government (MSIT) (No.2019-0-01906, Artificial Intelligence Graduate School Program(POSTECH), No.2024-00457882, AI Research Hub Project). This work was also partly supported by Samsung Research Funding Center (SRFC-IT1801-52).

%This work was supported by \rt{the National Research Foundation of Korea (NRF) grant funded by the Korea government (MSIT) (NRF-2018R1A5A1060031), Institute of Information \& communications Technology Planning \& Evaluation (IITP) grant funded by the Korea government (MSIT) (No.2019-0-01906, Artificial Intelligence Graduate School Program(POSTECH)), and Samsung Electronics Co., Ltd.}

\clearpage 

% ---- Bibliography ----
%
% BibTeX users should specify bibliography style 'splncs04'.
% References will then be sorted and formatted in the correct style.
%
\bibliographystyle{splncs04}
\bibliography{main}

\clearpage 

% ---------------------------------------------------------------
% TODO REVIEW: Replace with your title
\title{Diffusion Model Compression\\for Image-to-Image Translation\\--Supplemental Document--} 

% TODO REVIEW: If the paper title is too long for the running head, you can set
% an abbreviated paper title here. If not, comment out.
\titlerunning{ID-Compression}

% TODO FINAL: Replace with your author list. 
% Include the authors' OCRID for the camera-ready version, if at all possible.
\author{Geonung Kim \and
Beomsu Kim \and
Eunhyeok Park \and
Sunghyun Cho}

% TODO FINAL: Replace with an abbreviated list of authors.
\authorrunning{G.~Kim et al.}
% First names are abbreviated in the running head.
% If there are more than two authors, 'et al.' is used.

% TODO FINAL: Replace with your institution list.
\institute{POSTECH\\
\email{\{k2woong92,qjatn0120,eh.park,s.cho\}@postech.ac.kr}\\
\url{https://kimgeonung.github.io/id-compression}
}

\maketitle

\setlength{\intextsep}{5pt}
\setlength{\columnsep}{5pt}

\setcounter{section}{0}
\setcounter{figure}{0}
\setcounter{table}{0}

\renewcommand\thesection{S\arabic{section}}
\renewcommand\thefigure{S\arabic{figure}}
\renewcommand\thetable{S\arabic{table}}

\begin{figure}[t]
  \centering
  \includegraphics[width=0.86\columnwidth]{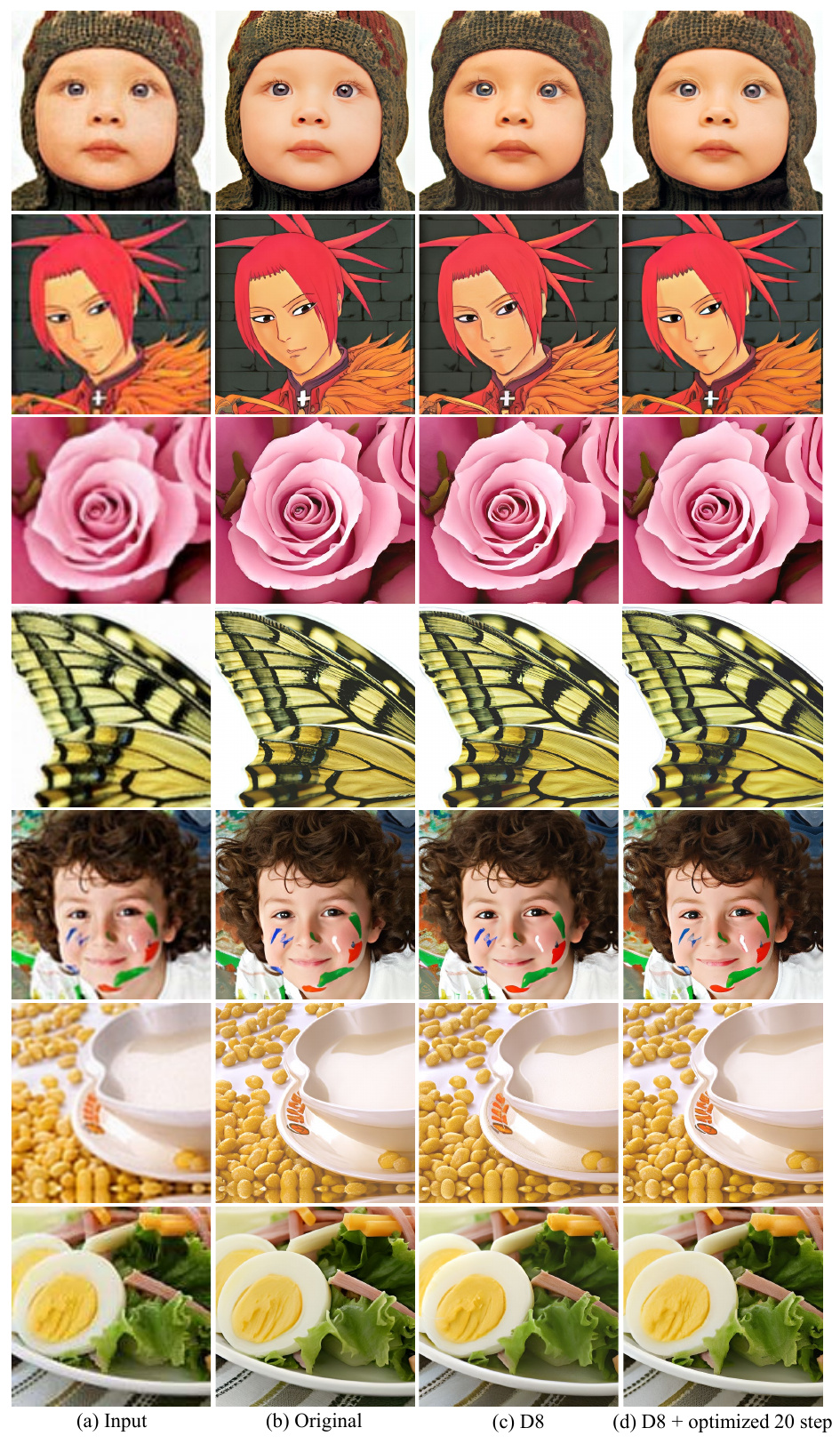}
  \vspace{-3mm}
  \caption{Additional qualitative results from StableSR~\cite{stablesr}. (c) shows the outputs after applying depth-skip compression at $D8$ with 50 steps. (d) displays the outcomes that combine depth-skip and time-step optimization.}
  \label{fig:supple-qual}
\end{figure}

In this Supplementary Material, we present:

\begin{itemize}
  \item Implementation details (\Sec{\ref{sec:detail}})
  \item Discussions on simple editing with shallower depth on IP2P (\Sec{\ref{sec:ds}})
  \item Additional analyses on the time-step optimization (\Sec{\ref{sec:ts-ablation}}) 
  % \item Comparisons and extensions with Diff-Pruning (\cref{sec:diff-pruning})
  \item Additional comparisons with AutoDiffusion (\Sec{\ref{sec:autodiff}})
  \item Additional details and results on applications (\Sec{\ref{sec:app}})
  \item Discussions on the multi-depth search of IP2P (\Sec{\ref{sec:multi-depth}})
  \item Profiling on the parameter number and latency for Stable Diffusion (\Sec{\ref{sec:profile}})
  \item Discussions on a challenge of step distillation in image restoration (\cref{sec:image-restoration})
  \item Additional latency and computational cost analysis (\cref{sec:ds-cost})
  \item Time-step optimization using different scheduler (\cref{sec:dpm-solver})
  \item Additional qualitative results of StableSR (\cref{fig:supple-qual})
\end{itemize}

\section{Implementation Details}
\label{sec:detail}
For the depth-search, we randomly sample 100 images from the training dataset of each task. In StableSR~\cite{stablesr}, we set the PSNR threshold at 28 dB, using the ground truth based on 50 iteration results. Regarding IP2P~\cite{ip2p}, we utilized a combination of CLIP image similarity~\cite{clip} and directional CLIP similarity~\cite{stylegannada}, which are the same evaluation protocol used in IP2P~\cite{ip2p}. The threshold values for these metrics were set at 0.7 and 0.2, respectively. 
Regarding ControlNet~\cite{controlnet}, we set the FID~\cite{fid} threshold at 22.

\section{Simple Editing with Shallower Depth on IP2P}
\label{sec:ds}

\begin{figure*}[t]
  \centering
  \includegraphics[width=1.0\linewidth]{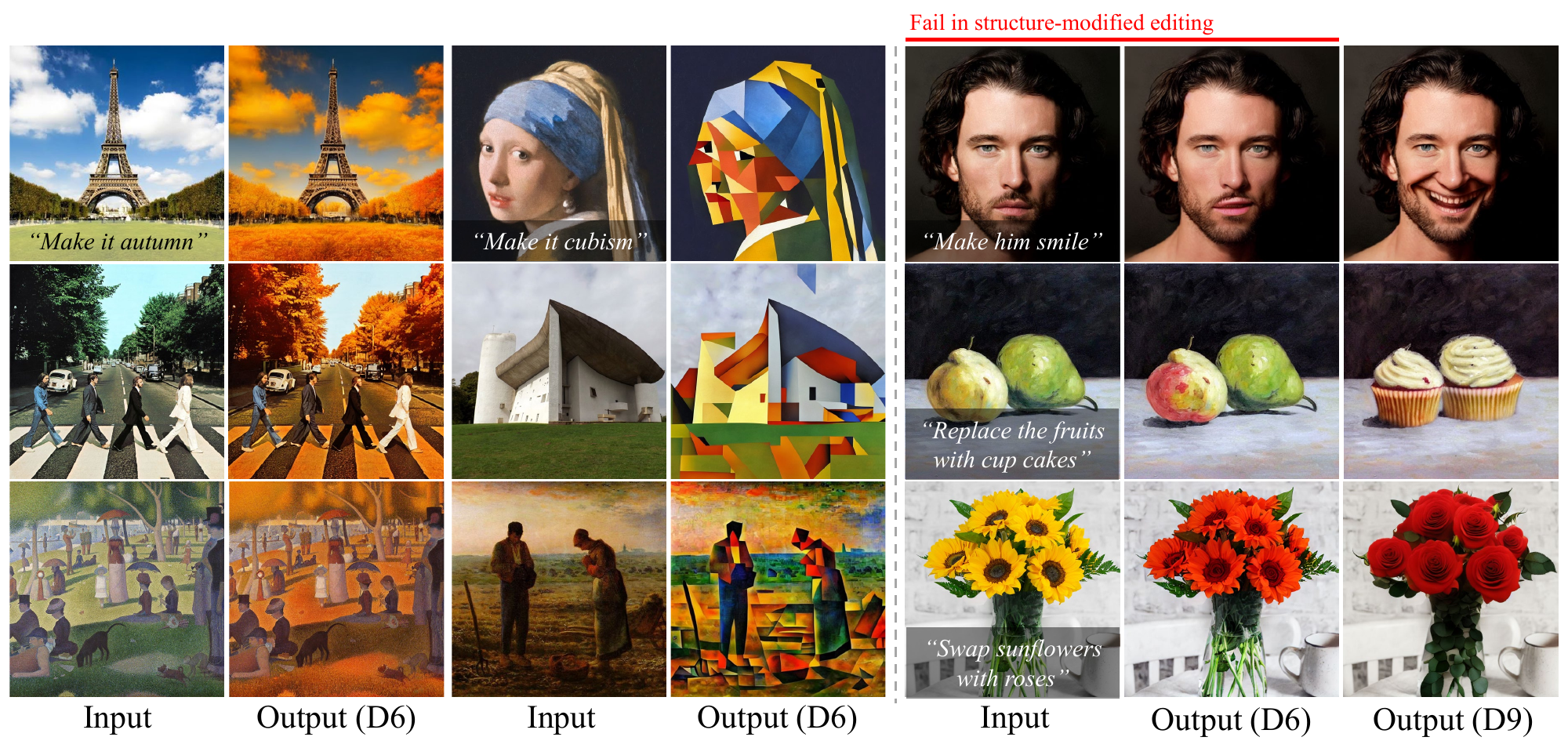}
  \caption{Various structure-preserved editing operates even after depth-skip compression at level 6, which uses 15.6\% parameters. 
The shallower the depth-skip, the more complex edits, such as structure-modified editing, tend to fail. 
  }
  \label{fig:d6}
\end{figure*}

% \begin{figure}[h]
%   \centering
%   \includegraphics[width=1.0\columnwidth]{assets/d6.pdf}
%   \caption{Various structure-preserved editings operate even after depth-skip compression at level 6, which uses 15.6\% parameters.
%   }
%   \label{fig:d6}
% \end{figure}

IP2P~\cite{ip2p} supports a wide range of image editing operations with various levels of difficulty.
This implies that different input text commands to IP2P may require different depth levels, i.e., easy commands that do not alter the image structure may require shallower depths than other commands.
\Fig{\ref{fig:d6}} shows examples of such easy and difficult commands.
In the figure, the commands ``Make it autumn'' and ``Make it cubism'' are easy ones that do not alter image structures, while the other commands shown on the right are difficult ones that change image structures.
As shown in the figure, the shallow-depth model with only six depth levels (D6 in the figure), which uses only 15.6\% parameters, successfully produces plausible results for the easy commands, but fails for the difficult ones, for which, the deeper-depth model with nine depth levels (D9) succeeds.
This result suggests that, by limiting target image editing operations to simple structure-preserving ones, more effective depth-skip compression can be achieved.

\section{Additional Analyses on Time-step Optimization}
\label{sec:ts-ablation}

\begin{figure}[t]
  \centering
  \includegraphics[width=1.0\columnwidth]{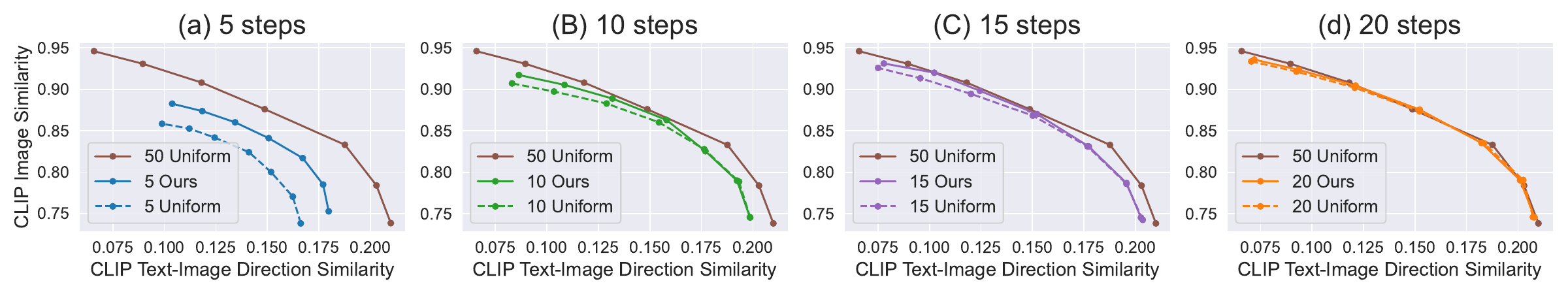}
  \caption{Quantitative comparison on time-step optimization for InstructPix2Pix~\cite{ip2p}.}
  \label{fig:ip2p-ts}
\end{figure}
 
\begin{table}[t]

\caption{The optimized values for the time-step control parameter $\gamma$ varies with change of CFG~\cite{cfg} values.} 

\centering
\scalebox{0.85}{

\begin{tabular}{ccccc}
\Xhline{2\arrayrulewidth}
\multirow{2}{*}{Iteration} & \multicolumn{4}{c}{$CFG_{I}$} \\
                           & 1.0   & 1.2   & 1.4   & 1.6   \\ \hline\hline
5                          & 0.322 & 0.333 & 0.377 & 0.435 \\
10                         & 0.556 & 0.581 & 0.617 & 0.704 \\
15                         & 0.709 & 0.714 & 0.833 & 0.833 \\ \Xhline{2\arrayrulewidth}
\end{tabular}

}

  \label{tbl:ts-cfg}
\end{table}

\begin{table}[t]

\caption{Time-step optimization results with and without depth-skip compression. The optimized values for the time-step control parameter $\gamma$ are almost identical between the models with and without depth-skip compression, indicating that the time-step optimization is not affected by the depth-skip compression.}

\centering
\scalebox{0.85}{

\begin{tabular}{ccccccc}
\Xhline{2\arrayrulewidth}
       & \multicolumn{2}{c}{5 step} & \multicolumn{2}{c}{10 step} & \multicolumn{2}{c}{15 step} \\ 
Depth-skip &  -       & \checkmark  &     -      & \checkmark   &     -      & \checkmark     \\ \hline\hline
InstructPix2Pix~\cite{ip2p}   & 0.322     & 0.322     & 0.556      & 0.556     & 0.709      & 0.714     \\
StableSR~\cite{stablesr} & 1.900     & 1.900        & 1.600      & 1.600        & 1.610      & 1.610       \\
\Xhline{2\arrayrulewidth}
\end{tabular}

}

  \label{tbl:ts-gamma}
\end{table}

\Fig{\ref{fig:ip2p-ts}} presents additional experimental results on our time-step optimization for various iteration numbers, conducted on IP2P~\cite{ip2p}.
Our time-step optimization achieves superior results compared to uniform sampling in terms of both CLIP image similarity and CLIP text-image direction similarity for small iteration numbers, as demonstrated in \Fig{\ref{fig:ip2p-ts}} (a).
The differences among the uniform sequence, our optimized sequence and 50-step sampling become marginal at 20 steps, as shown in \Fig{\ref{fig:ip2p-ts}} (d).
This suggests that by transferring to simpler downstream tasks, the necessary number of iterations is naturally reduced compared to the conventional 50-step approach.

\paragraph{Impact of CFG}
Classifier Free Guidance (CFG)~\cite{cfg} is a method that is widely used in numerous conditional diffusion models to control the influence of conditions on image generation.
CFG provides a single strength parameter for each condition to control its influence.
To obtain a high-quality result that a user wants, they often run diffusion models multiple times with different values for the CFG strength parameters.
IP2P~\cite{ip2p}, which is one of our target applications, also utilizes CFG to control the strength of prompt or image conditions. 

In this section, we additionally present an analysis on the impact of CFG on the time-step optimization.
Specifically, we conduct the time-step optimization of IP2P~\cite{ip2p} with different CFG strength parameter values for the image condition, and with different iteration numbers.
\Tbl{\ref{tbl:ts-cfg}} shows the result where $CFG_{I}$ indicates the CFG strength parameter for the image condition.
As the result reveals, changing the CFG parameter leads to significantly different values for the time-step control parameter $\gamma$, which indicate different optimal time step sequences, for all iteration numbers.

This result also validates the practicality of our computationally-efficient time-step optimization method.
To achieve high-quality results for different CFG parameter values requires to perform time-step optimization multiple times, which can be excessively time-consuming given the wide range of CFG parameter values.
Nevertheless, our time-step optimization can significantly reduce the computational overhead compared to AutoDiffusion~\cite{autodiff}, which is the previous state-of-the-art method, as our method performs at least 62 times faster than AutoDiffusion, as shown in Table 4 in the main paper.

\paragraph{Impact of Depth-skip on Time-step optimization}
We analyze the impact of depth-skip on the time-step optimization.
To this end, we prepare two diffusion models: one is an original full model, and the other is a depth-skipped version obtained using our depth-skip compression.
We then perform time-step optimization on both models, and compare the results.
\Tbl{\ref{tbl:ts-gamma}} shows the time-step optimization results, where the optimal values for the time-step control parameter $\gamma$ are not affected by depth skip.
This result indicates that the time-step optimization and depth-skip compression are independent to each other, allowing them to be performed parallel.

\begin{figure}[t!]
  \begin{center}
    \includegraphics[width=0.7\linewidth]{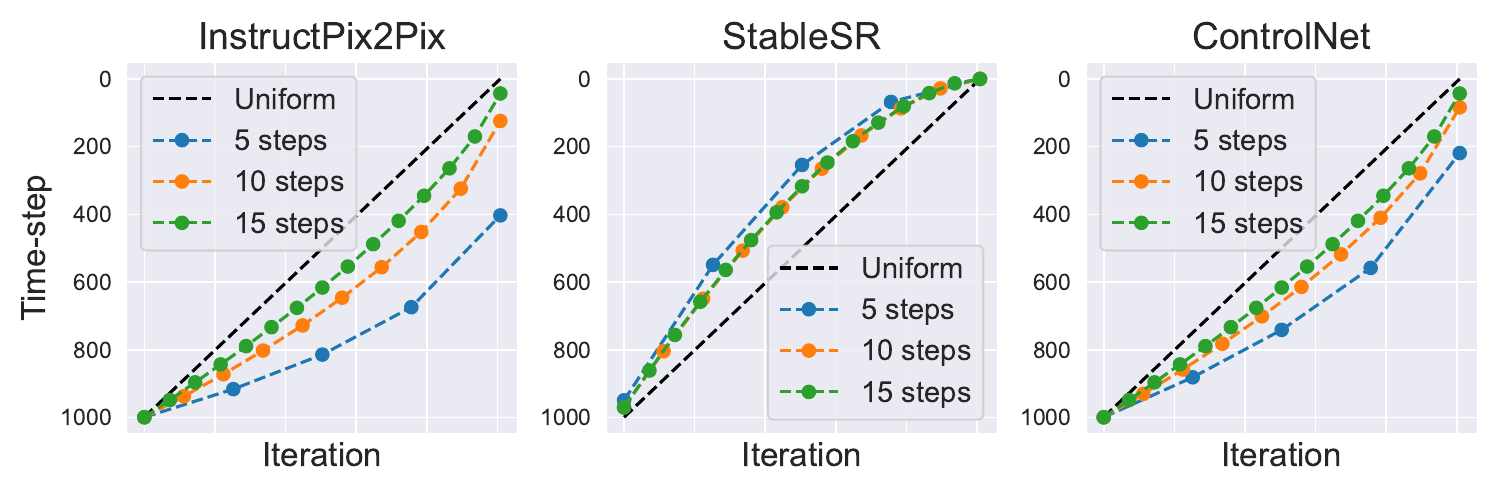}
  \end{center}
  \vspace{-7mm}
  \caption{Visualization for our optimized time-steps}
  \vspace{-3mm}
  \label{fig:vis}
\end{figure}

\paragraph{Optimized time-step visualization}
\cref{fig:vis} visualizes the results of our optimized time-steps. 
Consistent with our intuition, the results show that later time steps are crucial in StableSR, whereas early time steps are more important in other tasks.

\section{Comparisons on Time-step Optimization with AutoDiffusion} % \fix{정성적으로 차이가 미비해서 예시를 보일 수 없음} 
\label{sec:autodiff}
\begin{table}[t]
\centering
\caption{
Quantitative comparison with AutoDiffusion for ControlNet~\cite{controlnet}.
}

\scalebox{0.75}{
\begin{tabular}{ccccc}
\Xhline{2\arrayrulewidth}
Step & Method   & FID$\downarrow$  & CLIP-Score$\uparrow$   & CLIP-a $\uparrow$  \\
\hline\hline
% 5    & Xue et al.~\cite{xue2024accelerating}      & 46.19            & 29.65                  & 5.39               \\
5    & AutoDiffusion~\cite{autodiff} & 28.77            & 30.35                  & 5.81               \\
5    & Ours     & \textbf{28.26}   & \textbf{30.38}         & \textbf{5.85}      \\
\hline
% 10   & Xue et al.~\cite{xue2024accelerating}      & 29.58            & 29.94                  & 5.77               \\
10   & AutoDiffusion~\cite{autodiff} & 24.21            & 30.39                  & 5.98               \\
10   & Ours     & \textbf{23.29}   & \textbf{30.40}         & \textbf{6.00}      \\
\Xhline{2\arrayrulewidth}
\label{tbl:ts-controlnet}
\end{tabular}
}

\end{table}
\begin{table*}[t]
    \begin{minipage}[b]{0.48\textwidth}

        \centering
        
        \caption{Comparison with AutoDiffusion for InstructPix2Pix~\cite{ip2p} at 5 steps.}
        
        \scalebox{0.75}{
        \begin{tabular}{cccl}
        \Xhline{2\arrayrulewidth}
        Method        & $\operatorname{CLIP}_{I}$ & $\operatorname{CLIP}_{D}$ & Search Time        \\ \hline\hline
        Ours          & \textbf{0.767}            & \textbf{0.190}              & \textbf{38.7 min} \\
        AutoDiffusion & 0.765                     & 0.188                       & 40.5 hour         \\ \Xhline{2\arrayrulewidth}
        \end{tabular}
        }
        \label{tbl:autodiff-ip2p}

    \end{minipage}
    \begin{minipage}[b]{0.48\textwidth}
        % \raggedright 
        \caption{Comparison with AutoDiffusion for StableSR~\cite{stablesr} at 10 steps.}
        \scalebox{0.75}{
        
        \begin{tabular}{ccccl}
        \Xhline{2\arrayrulewidth}
        Method        & FID             & PSNR            & LPIPS          & Search Time        \\ \hline\hline
        Ours          & \textbf{34.956} & 22.250          & \textbf{0.461} & \textbf{11.1 min} \\
        AutoDiffusion & 36.496          & \textbf{22.347} & 0.466          & 27.1 hour          \\ \Xhline{2\arrayrulewidth}      
        \end{tabular}
        }
        \label{tbl:autodiff-sdsr}
        
    \end{minipage}
    \vspace{-0mm}
\end{table*}

In this section, we present additional quantitative comparisons on time-step optimization with AutoDiffusion~\cite{autodiff} using IP2P~\cite{ip2p}, StableSR~\cite{stablesr} and ControlNet~\cite{controlnet}.
Regarding IP2P,
we set the CFG strength parameters for the image and text conditions to 1.0 and 7.5, respectively,
and optimize its time steps using our time-step optimization approach and AutoDiffusion for five iterations.
We then evaluate the time-step optimization results using the CLIP image similarity~\cite{clip} and directional CLIP similarity~\cite{stylegannada}.
\Tbl{\ref{tbl:autodiff-ip2p}} shows a comparison between our result and the result of AutoDiffusion.
As the table shows, our method achieves higher CLIP image similarity and directional CLIP similarity scores within a much shorter search time, proving the effectiveness of our approach.

For comparison on StableSR~\cite{stablesr}, we optimize the time steps of StableSR for 10 iterations using our time-step optimization approach and AutoDiffusion~\cite{autodiff}. We then evaluate their results using the DIV2K validation dataset~\cite{div2k}. \Tbl{\ref{tbl:autodiff-sdsr}} shows that our approach achieves comparable results in terms of FID~\cite{fid}, PSNR, and LPIPS~\cite{lpips} within a significantly shorter search time.

Regarding ControlNet~\cite{stablesr}, we conduct comparisons for 5 and 10 iterations. We then evaluate the results using the 5K COCO~\cite{coco} validation dataset. \Tbl{\ref{tbl:ts-controlnet}} shows that our approach outperforms for all criteria.

\section{Additional Applications}
\label{sec:app}

\begin{figure*}[t]
  \centering
  \includegraphics[width=1.0\linewidth]{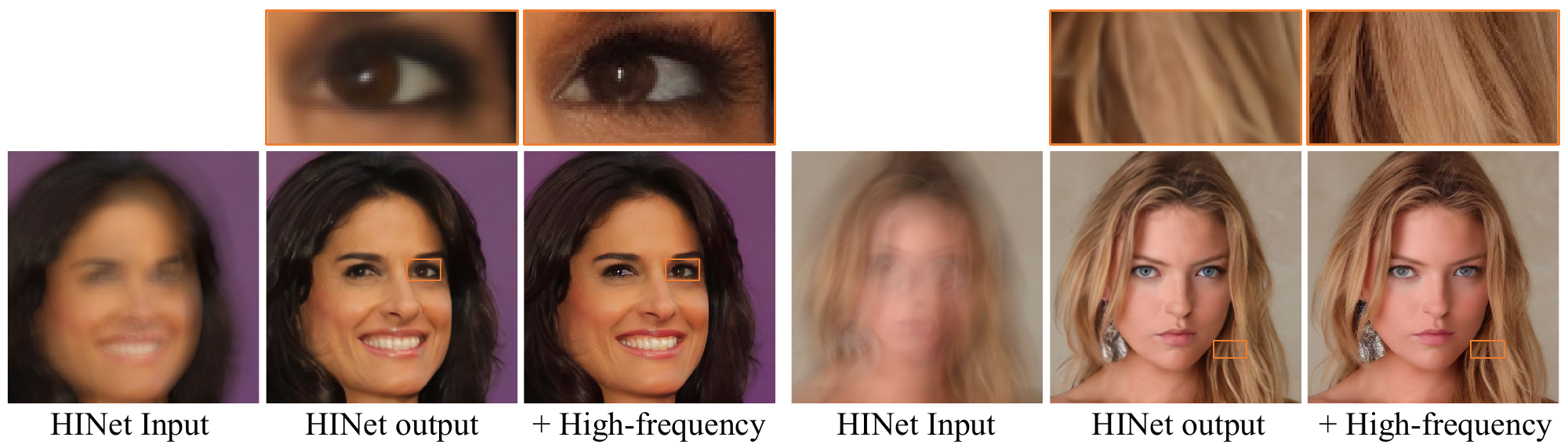}
  \caption{Qualitative results on high-frequency synthesis application.}
  \label{fig:high-freq}
\end{figure*}

\begin{figure*}[t]
  \centering
  \includegraphics[width=1.0\linewidth]{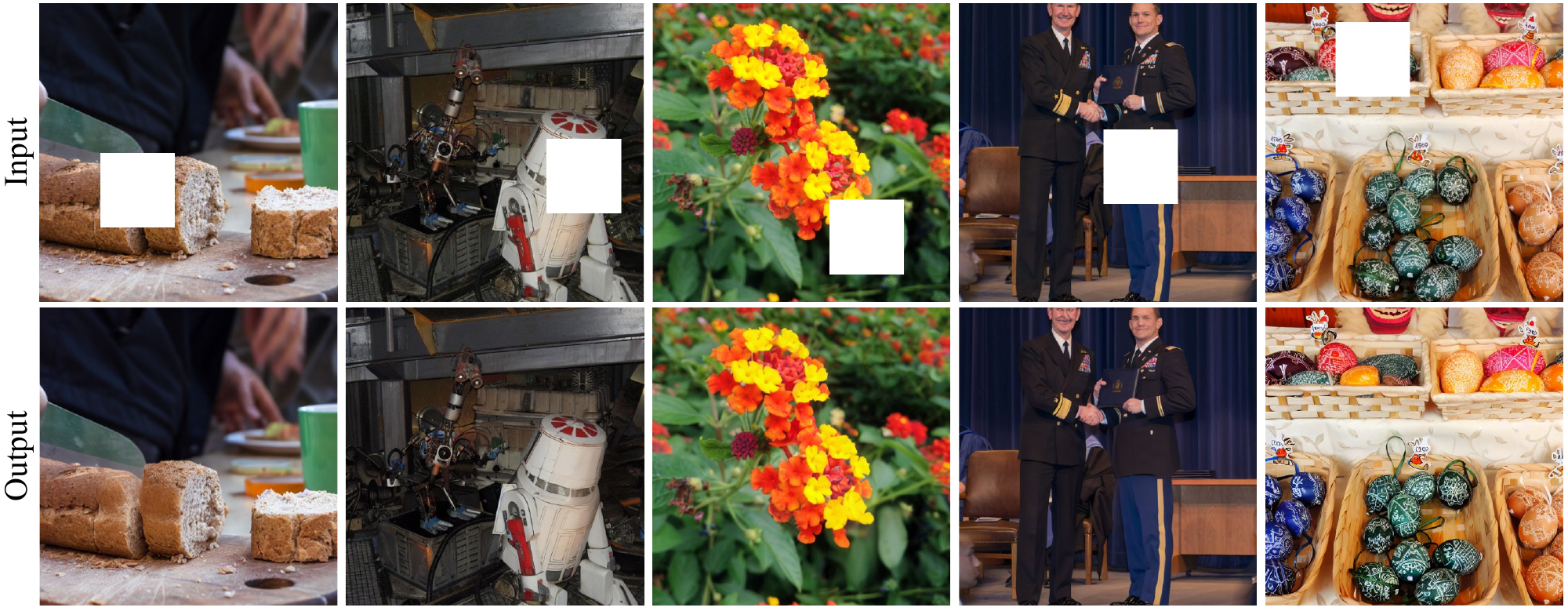}
  \caption{Qualitative results on image inpainting application.}
  \label{fig:inpainting}
\end{figure*}

% \subsection{Applications}
% \label{exp:app}

% To show a versatility of our method, we apply our method to high-frequency synthesis and image inpainting tasks.

% \vspace{-3mm}
% \paragraph{High-frequency synthesis} 
% A recent approach to image restoration involves combining a regression approach with a generative prior method for faithful restoration with natural-looking textures~\cite{reconandgen,ugpnet}.
% %A recent approach to image restoration involves combining a regression approach, which is weak in handling high-frequency details, with a generative prior method, which may lack content consistency~\cite{reconandgen,ugpnet}.
% In this application, our method can offer a resource-efficient generative prior. For transfer learning, we utilize Stable Diffusion~\cite{ldm} as upstream model. \Fig{\ref{fig:app}} (a) demonstrates that our model successfully generates high-frequency details over images restored by HINet~\cite{hinet}, using only 26.6\% of the parameters and 20 steps.

% \vspace{-3mm}
% \paragraph{Image inpainting}
% We adapt a text-conditioned image inpainting model~\cite{ldm} for unconditional inpainting tasks. \Fig{\ref{fig:app}} (b) illustrates that our model can generate plausible content, utilizing 43.3\% of the parameters and 15 steps.

To train the high-frequency synthesis model in Sec.~4.5 in the main paper, we use paired images consisting of a restored image and its ground-truth. 
To collect the restored images, we employ a pretrained HINet~\cite{hinet}, trained for deblurring using the CelebA dataset~\cite{celeba}.
To accommodate the additional input image, we implement minor modifications to the input network layer, same as IP2P~\cite{ip2p}.
To train the image inpainting model, we synthesize paired images where each image pair consists of a masked-image and its ground-truth image using the OpenImage dataset~\cite{openimage}.
The text embedding is fixed as an empty text.
\Fig{\ref{fig:high-freq}} and \Fig{\ref{fig:inpainting}} shows additional qualitative examples of the high-frequency synthesis and image inpainting models, respectively.

\section{Analysis on Multi-depth Search on IP2P}
\label{sec:multi-depth}

\begin{table*}[t]
\centering
\caption{Comparisons between single and multi-depth search applied to InstructPix2Pix~\cite{ip2p}.
The percentages at ``Time'' and ``Param'', which indicates parameter, are proportion to requirements in full model.
This results suggest that our single-depth approach efficiently searches the near-optimal point compared to a multi-depth search strategy. 
}
\scalebox{0.7}{
\begin{tabular}{c|c|cc|cc|cc}
\Xhline{2\arrayrulewidth}
\multicolumn{1}{l|}{} & (a) Fix memory \& Min time & \multicolumn{2}{c|}{(b) Fix time \& Max $\operatorname{CLIP}_I$} & \multicolumn{2}{c|}{(c) Fix time \& Max $\operatorname{CLIP}_{D}$} & \multicolumn{2}{c}{(d) Fix quality \& Min time}          \\
\multicolumn{1}{l|}{} & ( $\Delta \operatorname{CLIP} <$ 0.005 )   & \multicolumn{2}{c|}{( $\Delta$Time $<$ 1\% )}    & \multicolumn{2}{c|}{( $\Delta$Time $<$ 1\% )}    & \multicolumn{2}{c}{( $\Delta \operatorname{CLIP} <$ 0.005 )} \\ 
Depth                & $\Delta$Time(\%)           & $\Delta \operatorname{CLIP}_I$        & $\Delta$Param(\%)       & $\Delta \operatorname{CLIP}_{D}$          & $\Delta$Param(\%)          & $\Delta$Time(\%)                & $\Delta$Param(\%)               \\ \hline\hline
%11                   & -10.83                     & +0.00                 & +9.37                   & +0.01                   & +9.37                 & -10.85                      & -9.37                      \\
10                   & -7.46                      & +0.04                 & +18.73                  & +0.01                   & +18.73                & -7.46                       & +0.00                       \\
9                    & -2.82                      & +0.03                 & +27.90                  & +0.01                   & +18.54                & -6.30                       & +18.54                      \\
8                    & -6.64                      & +0.04                 & +36.01                  & +0.07                   & +26.64                & -10.46                      & +17.47                      \\\Xhline{2\arrayrulewidth}
\end{tabular}
}

\label{tbl:multi-depth}
\end{table*}

We demonstrated the efficiency of our single-depth search over multi-depth search on StableSR~\cite{stablesr} in Sec. 3.3 in the main paper.
This section presents a similar analysis on IP2P~\cite{ip2p}, which is detailed in \Tbl{\ref{tbl:multi-depth}}.
we use the CLIP image similarity~\cite{clip} and directional CLIP similarity~\cite{stylegannada} for quality assessment.
When the model size is constrained, the multi-depth search solution achieves marginal latency improvement over ours, while maintaining comparable quality, as shown in \Tbl{\ref{tbl:multi-depth}} (a).
When the latency is constrained, optimizing the quality using the multi-depth search leads to a significant increase in the parameter number, as shown in \Tbl{\ref{tbl:multi-depth}} (b) and (c).
When the quality is constrained, minimizing the latency also results in a trade-off involving an increase in the parameter number, as shown in \Tbl{\ref{tbl:multi-depth}} (d).
All these results suggest that our single-depth search is already able to achieve near-optimal solutions despite its considerably smaller search space.

\section{Profiling of Stable Diffusion}
\label{sec:profile}

\begin{table}[t]

\caption{Parameter number and latency for each depth level to Stable Diffusion (version 1). Half of the deeper depth levels occupy 73.4\% of the parameters.} 

\centering
\scalebox{0.85}{

\begin{tabular}{c|rr|rr}
\Xhline{2\arrayrulewidth}
Depth & \multicolumn{1}{c}{Parameter} & \multicolumn{1}{c|}{Sum} & \multicolumn{1}{c}{Latency} & \multicolumn{1}{c}{Sum} \\
\hline\hline
D01   & 0.69\%                        & 0.70\%                  & 5.09\%                      & 5.09\%                  \\
D02   & 1.25\%                        & 1.95\%                  & 10.65\%                     & 15.74\%                 \\
D03   & 1.37\%                        & 3.32\%                  & 10.77\%                     & 26.51\%                 \\
D04   & 2.85\%                        & 6.17\%                  & 8.57\%                      & 35.08\%                 \\
D05   & 4.39\%                        & 10.56\%                 & 13.03\%                     & 48.10\%                 \\
D06   & 5.06\%                        & 15.62\%                 & 13.06\%                     & 61.16\%                 \\
D07   & 10.98\%                       & 26.60\%                 & 5.06\%                      & 66.22\%                 \\
D08   & 16.71\%                       & 43.31\%                 & 10.49\%                     & 76.70\%                 \\
D09   & 17.47\%                       & 60.78\%                 & 9.22\%                      & 85.93\%                 \\
D10   & 9.17\%                        & 69.95\%                 & 1.88\%                      & 87.80\%                 \\
D11   & 9.37\%                        & 79.32\%                 & 2.76\%                      & 90.56\%                 \\
D12   & 9.37\%                        & 88.68\%                 & 2.70\%                      & 93.26\%                 \\
D13   & 11.32\%                       & 100.0\%                 & 6.74\%                      & 100.00\%                \\
\Xhline{2\arrayrulewidth}
\end{tabular}
}

  \label{tbl:profile-sd1}
\end{table}

\Tbl{\ref{tbl:profile-sd1}} shows the proportions of the parameter number and latency of each depth level.
It is important to note that half of the deeper depth levels occupy 73.4\% of the parameters. 
This is because the layers at coarser levels tend to have more channels, which contributes to an increase in the number of parameters. 
This tendency enhances the effectiveness of the depth-skip compression method in reducing the number of parameters, as our approach begins by discarding the coarsest layers first.

\section{Challenge of Step Distillation in Image Restoration}
\label{sec:image-restoration}

\begin{figure}[t]
  \centering
  \includegraphics[width=1.0\columnwidth]{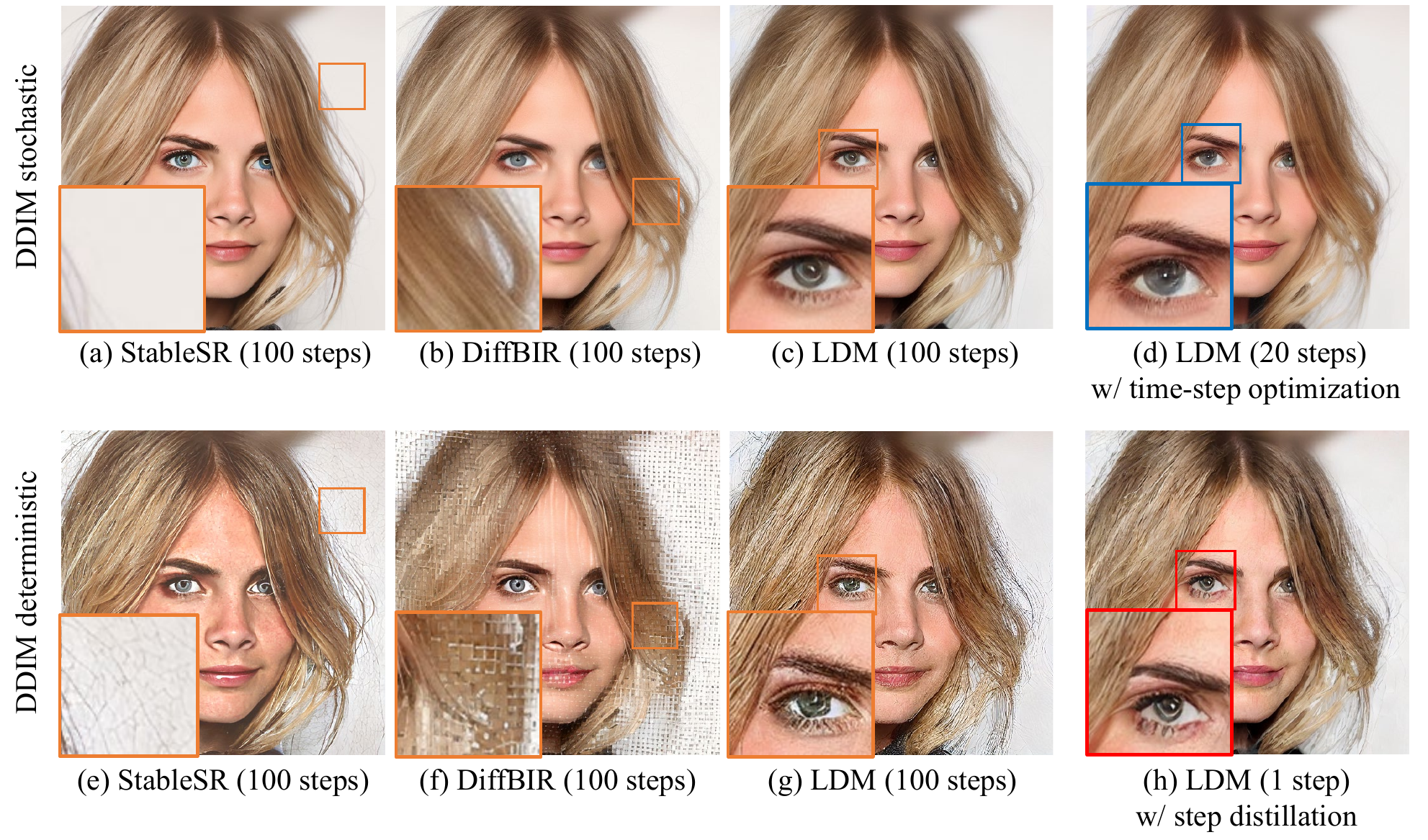}
  % \caption{Super-resolution outcomes using DDIM stochastic and deterministic process across various models, such as StableSR~\cite{stablesr}, DiffBIR~\cite{diffbir}, and LDM~\cite{ldm}.}
  \caption{Super-resolution outcomes using various configurations. Top and bottom row are outcomes using stochastic and deterministic sampling, respectively. (d) is a sampling result using our time-step optimization. (h) is a sampling result after applying a step distillation method~\cite{vprediction}.}
  \vspace{-1mm}
  \label{fig:restoration}
\end{figure}

% 흐름 역전
Step distillation \cite{vprediction,ondistillation,add,instaflow,lcm,snapfusion} is an acceleration technique capable of more aggressively speeding up diffusion models compared to our method, potentially reducing the diffusion process to only one or two iterations.
Thus, one may want to adopt step distillation to a downstream task such as image restoration.
However, unlike our training-free acceleration method, step distillation techniques are not suited for image restoration tasks.
In this section, we analyze the the reason behind such incompatibility of step distillation, contrasting it with the effectiveness of our time-step optimization.

Step distillation methods alter the objective of diffusion models from functioning as progressive denoisers to simply approximating the posterior mean. 
This transformation compels diffusion models to adhere to the deterministic process for sampling.
However, this deterministic sampling in image restoration results in artifact-prone outcomes.
\cref{fig:restoration} exemplifies this issue, where (a-c) and (e-g) illustrate restoration results across various models, such as StableSR~\cite{stablesr}, DiffBIR~\cite{diffbir}, and LDM~\cite{ldm}, with 100 steps using the stochastic and deterministic processes of DDIM~\cite{ddim}, respectively.
While the stochastic sampling consistently yields higher-quality outputs, the deterministic process frequently results in artifacts, including unnatural textures and noisy high-frequency details. 
This issue is inevitably reproduced in step-distilled models, as shown in \cref{fig:restoration} (h), since they should use the deterministic sampling process.
In contrast, our time-step optimization adopts a training-free acceleration approach, ensuring broad compatibility with downstream applications by preserving the original models and their sampling processes, as shown in \cref{fig:restoration} (d).

A main reason of the failure of the deterministic sampling may be attributed to the local optimum problem.
Specifically, the output of diffusion models represents the gradient of the data distribution at a time step, denoted as $\nabla_x \log p_t(x)$~\cite{sde}.
Stochastic sampling employs this gradient in an iterative process that subtracts the gradient while simultaneously introducing random noise.
This process can be interpreted as gradient descent with simulated annealing, facilitating a more comprehensive exploration of the solution space.
Meanwhile, the deterministic process removes the stochastic component, functioning akin to the pure gradient decent method. 
This modification suggests that without stochastic perturbations, there is an elevated risk becoming trapped in local optima with unnatural textures.

\begin{table}[t!]

\caption{
Latency and MACs~\cite{thop} improvement after depth-skip. We measure the latencies and computational costs with 50 iterations for IP2P~\cite{ip2p} and StableSR~\cite{stablesr}, and 20 iterations for ControlNet~\cite{controlnet}.
}

\centering
\scalebox{0.85}{

\begin{tabular}{ccccccc}
\Xhline{2\arrayrulewidth}
\multicolumn{1}{l}{} & \multicolumn{2}{c}{InstructPix2Pix~\cite{ip2p}} & \multicolumn{2}{c}{StableSR~\cite{stablesr}} & \multicolumn{2}{c}{ControlNet~\cite{controlnet}} \\
\multicolumn{1}{l}{} & Time(s)          & MACs(T)          & Time(s)       & MACs(T)      & Time(s)        & MACs(T)       \\ \hline\hline
Original             & 6.312            & 50.81            & 2.807         & 18.75        & 1.572          & 18.21         \\
D9                   & 5.978            & 47.27            & 2.421         & 17.49        & 1.444          & 17.26         \\
D8                   & 5.685            & 41.75            & 2.204         & 15.57        & 1.372          & 15.79         \\
        \Xhline{2\arrayrulewidth}

\end{tabular}
}

  \label{tbl:time-quant1}
\end{table}

\section{Latency \& Computational Cost Analysis}
\label{sec:ds-cost}
\cref{tbl:time-quant1} reports the reduction of latencies and computational costs achieved by applying depth-skip pruning.
The table shows that depth-skip pruning reduces the latency and computation of 5.3\% and 7.0\% for IP2P~\cite{ip2p}, 21.5\% and 17.0\% for StableSR~\cite{stablesr}, and 8.2\% and 5.2\% for ControlNet~\cite{controlnet}, respectively.
% \cref{tbl:timeconsum} shows the latency reduction achieved by applying depth-skip pruning and time-step optimization, which includes the variational auto-encoder (VAE) and text-encoder.

\begin{table}[t]
\centering
\caption{Quantitative comparisons of time-step optimization using multistep DPM-Solver++~\cite{dpm-solver} schedular. The evaluation protocol is same as the one used in Tab. 3 in the main paper. }
\scalebox{0.85}{
\begin{tabular}{cccccc}
        \Xhline{2\arrayrulewidth}
                         & \multicolumn{1}{l}{} & \multicolumn{2}{c}{InstructPix2Pix}                                     & \multicolumn{2}{c}{StableSR}                                 \\
                         & \# Steps             & \multicolumn{1}{c}{Ours} & \multicolumn{1}{c}{AutoDiffusion} & \multicolumn{1}{c}{Ours} & \multicolumn{1}{c}{AutoDiffusion} \\ \hline\hline
\multicolumn{1}{c}{PSNR} & 5                    & \textbf{20.54}           &  18.73                            &  \textbf{27.41}           &  26.79                           \\
\multicolumn{1}{c}{(dB)} & 10                   & \textbf{26.82}           &  24.33                            &  \textbf{32.09}           &  30.58                          \\
        \Xhline{2\arrayrulewidth}
\end{tabular}
}

\label{tbl:dpmsolver}
\end{table}

\section{Time-step optimization using different scheduler}
\label{sec:dpm-solver}
Although we evaluated our time-step optimization using DDIM sampler in the main paper, our method can be applied to other schedulers as well, since the optimization process imposes no constraints, such as differentiability. Also, it is still effective in other schedulers, such as multistep DPM-Solver++~\cite{dpm-solver}, as demonstrated in \ref{tbl:dpmsolver}.

\end{document}